\documentclass{article} 
\usepackage{iclr2026_conference,times}

 \usepackage{tikz}
 \usetikzlibrary{arrows.meta,positioning,fit,calc}

\usepackage{amsmath,amsfonts,bm}

\def\eqref#1{equation~\ref{#1}}

\def\1{\bm{1}}

\DeclareMathAlphabet{\mathsfit}{\encodingdefault}{\sfdefault}{m}{sl}
\SetMathAlphabet{\mathsfit}{bold}{\encodingdefault}{\sfdefault}{bx}{n}

\usepackage{hyperref}
\usepackage{url}
\usepackage{wrapfig}

\usepackage{booktabs}
\usepackage{longtable}
\usepackage{geometry}
\usepackage{xcolor}
\usepackage{subcaption}
\usepackage{supertabular}
\usepackage{titletoc}
\usepackage{float}
\usepackage{hyperref}

\usepackage{enumitem}

\definecolor{rightblue}{RGB}{76,114,176} 
\definecolor{rightorange}{RGB}{221,132,82} 
\definecolor{aliceblue}{rgb}{0.94, 0.97, 1.0} 
\definecolor{darkcerulean}{rgb}{0.03, 0.27, 0.49} 
\definecolor{iris}{rgb}{0.35, 0.31, 0.81} 
\definecolor{carmine}{rgb}{0.59, 0.0, 0.09} 
\definecolor{green(munsell)}{rgb}{0.0, 0.66, 0.47} 
\definecolor{springgreen}{RGB}{0,160,75}
\definecolor{celadon}{rgb}{0.67, 0.88, 0.69} 
\definecolor{bluerow}{rgb}{0.0, 0.53, 0.74} 
\definecolor{lightorange}{RGB}{255, 219, 187} 
\definecolor{lavenderblue}{rgb}{0.8, 0.8, 1.0}
\definecolor{blue(pigment)}{rgb}{0.2, 0.2, 0.6}
\definecolor{blue-violet}{rgb}{0.54, 0.17, 0.89}
\definecolor{seagreen}{RGB}{0,150,120}
\definecolor{blueseaborn}{RGB}{1,115,178}
\definecolor{orangeseaborn}{RGB}{222,143,6}
\definecolor{greenseaborn}{RGB}{1,158,115}

\usepackage{hyperref}
   \hypersetup{
     final=true,
     plainpages=false,
     pdfstartview=FitV,
     pdftoolbar=true,
     pdfmenubar=true,
     bookmarksopen=true,
     bookmarksnumbered=true,
     breaklinks=true,
     linktocpage=true,
     colorlinks=true,
     linkcolor=springgreen, 
     urlcolor=blueseaborn, 
     citecolor=darkcerulean, 
     anchorcolor=black
}

\title{\raisebox{-0.15ex}{\includegraphics[height=1em]{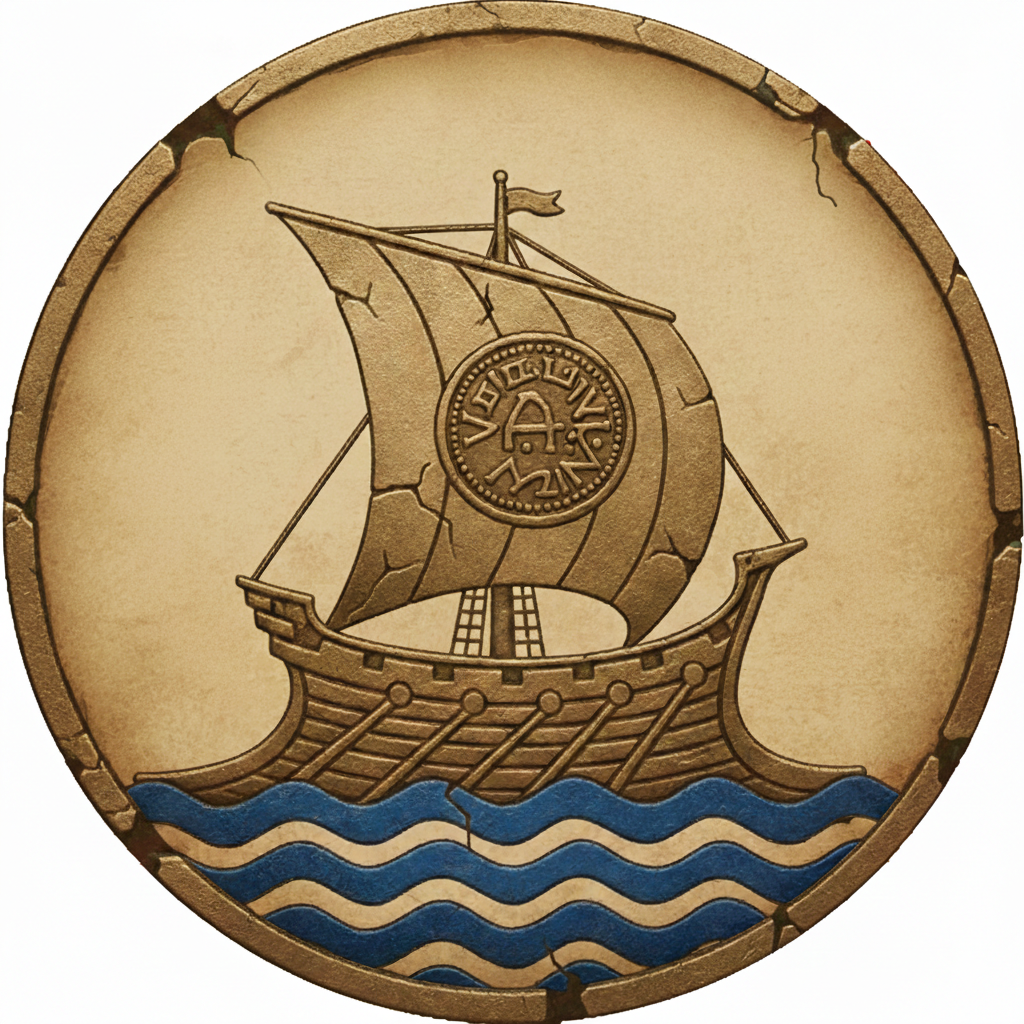}} UTICA: multi-objective self-distllation Foundation model pretraining for time series classification}

\iclrfinalcopy
\author{Yessin Moakher \\
Ecole Polytechnique, France\\
\textit{yessin.moakher@polytechnique.edu} \\
\And
Youssef Attia El Hili \\
Huawei Noah's Ark Lab
\And
Vasilii Feofanov \\
42.com
}

\track{Research}
\begin{document}

\maketitle
\begin{abstract}

Self-supervised foundation models have achieved remarkable success across domains, including time series. However, the potential of non-contrastive methods, a paradigm that has driven significant advances in computer vision, remains underexplored for time series. In this work, we adapt DINOv2-style self-distillation to pretrain a time series foundation model, building on the Mantis tokenizer and transformer encoder architecture as our backbone. Through a student–teacher framework, our method Utica learns representations that capture both temporal invariance via augmented crops and fine-grained local structure via patch masking. Our approach achieves state-of-the-art classification performance on both UCR and UEA benchmarks. These results suggest that non-contrastive methods are a promising and complementary pretraining strategy for time series foundation models. \footnote{The code is available at: \url{https://www.github.com/fegounna/Utica}.}

\end{abstract}

\section{Introduction}

Foundation models have fundamentally reshaped machine learning. Building on their success in language and vision, researchers are now developing Time Series Foundation Models (TSFMs) to learn universal representations from temporal data. Most TSFMs focus on forecasting, employing autoregressive \citep{ansari2024chronoslearninglanguagetime}, supervised \citep{auer2025tirexzeroshotforecastinglong}, or masked reconstruction objectives \citep{goswami2024momentfamilyopentimeseries}. While effective for prediction, such objectives prioritize local temporal consistency over global semantic structure, which is vital for classification tasks such as fault detection \citep{zhou2025batterybertrealisticbatteryfault}, cardiovascular diagnostics \citep{li2025mira}, and EEG decoding \citep{gnassounou2025leveraginggenerictimeseries}.

One prevailing approach to bridging this gap is contrastive learning. Drawing inspiration from computer vision \citep{chen2020simple,radford2021learningtransferablevisualmodels}, TSFMs such as Mantis \citep{feofanov2025mantislightweightcalibratedfoundation} have achieved remarkable performance using a contrastive objective: pulling together positive pairs (different augmentations of the same example) while pushing apart negative ones (different examples within a batch). However, this approach rests on the risky assumption that different samples within a batch are semantically distinct. In time series, where samples may share similar dynamics, frequency content, or temporal structure, this assumption often fails. This introduces false negatives, potentially harming representation quality and discouraging the model from capturing globally shared patterns.


Other approaches rely on self-distillation. \cite{pieper2023selfdistilledrepresentationlearningtime} propose a CNN-based student--teacher framework trained on masked views. Subsequently, NuTime \citep{lin2024nutimenumericallymultiscaledembedding} introduced a Transformer architecture trained with a BYOL-style \citep{NEURIPS2020_f3ada80d} self-distillation loss on pairs of randomly cropped global views.

While both methods avoid explicit negatives, they rely on a single view-generation strategy: masking only \citep{pieper2023selfdistilledrepresentationlearningtime} or paired global crops only \citep{lin2024nutimenumericallymultiscaledembedding}. In contrast, inspired by the success of DINOv2 \citep{oquab2024dinov2learningrobustvisual} in computer vision, we propose \textbf{Utica}, which pretrains a Transformer-based TSFM with a combination of (i) a self-distillation loss applied to heterogeneous global and local multi-crop augmentations, and (ii) a masking objective applied to the original sequence. We argue that combining masking with multi-crop augmentations is a natural fit for temporal data, facilitating the learning of global representations invariant to scale, partial observability and temporal offsets. We experimentally show the superiority of Utica on two time series benchmarks: UCR \citep{dau2019ucrtimeseriesarchive} and UEA \citep{bagnall2018ueamultivariatetimeseries}.


\section{Methodology}
\begin{wrapfigure}[12]{r}{0.4\textwidth}
    \centering
    \vspace{-0.5cm}\includegraphics[width=0.98\linewidth]{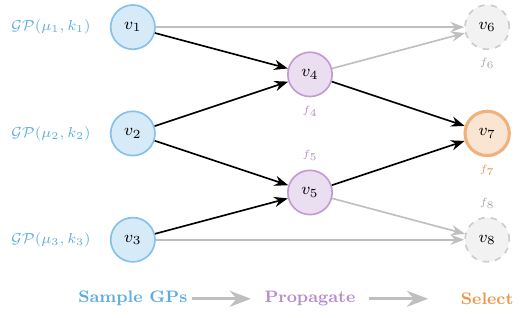}
    \caption{Example of a time series sample generation via causal DAG.}
    \label{fig:placeholder}
\end{wrapfigure}
\paragraph{Pretraining Dataset.} Following \cite{cauker2025} who showed that time series foundation models can be efficiently pretrained entirely on synthetic data, we generate synthetic sequences using a causal generative model defined by a directed acyclic graph (DAG). For each root node $v$, we sample a time series $x_v \in \mathbb{R}^T$ from a Gaussian Process
$x_v(t) \sim \mathcal{GP}(\mu_v(t), k_v(t,t'))$, where $k_v$ is a randomly composed covariance kernel
and $\mu_v$ is a non-stationary mean function.
For each non-root node $v$ with parents $\mathrm{pa}(v)$, the signal is generated as
$x_v(t) = f_v\!\left(\sum_{u \in \mathrm{pa}(v)} w_{uv} x_u(t) + b_v\right),$
with $w_{uv}, b_v \sim \mathcal{N}(0,1)$ and $f_v$ a randomly sampled nonlinearity.
A subset of nodes is selected to generate time series examples used for pretraining .
\begin{wrapfigure}[13]{l}
{0.4\textwidth}
    \centering
    \vspace{-0.5cm}\includegraphics[width=0.9\linewidth]{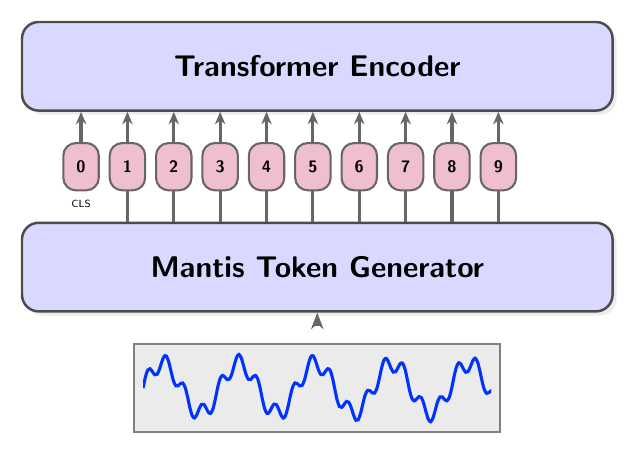}
    \caption{Architecture.}
\end{wrapfigure}
\textbf{Architecture. } As a backbone, we use a classical Transformer \citep{dosovitskiy2021imageworth16x16words} with a modality-specific token generator from Mantis \citep{feofanov2025mantislightweightcalibratedfoundation}. Each univariate input series is represented using three complementary transformations: the instance-normalized series, its first-order differential (to capture stationarity), and patch-level encodings of the mean and standard deviation of raw segments \citep{lin2024nutimenumericallymultiscaledembedding}. These embeddings are concatenated, projected to the model dimension $D=256$, and processed through 6 Transformer encoder layers, with a learnable class [CLS] token and sinusoidal positional encodings to preserve temporal information. The output that corresponds to the [CLS] token are treated as final embeddings produced by the model.
\textbf{Pretraining Loss.} The training setup of Utica consists of a Student network $f_{\theta_s}$ and a Teacher network $f_{\theta_t}$ with identical architectures. The Teacher weights are updated via an exponential moving average of the Student weights, following the schedule $\theta_t \leftarrow \lambda \theta_t + (1 - \lambda) \theta_s$, where $\lambda\in[0,1]$ increases linearly during pretraining. The total loss $\mathcal{L}$ is a sum of three distinct objectives:
$    \mathcal{L} = \mathcal{L}_{\text{DINO}} + \mathcal{L}_{\text{iBOT}} + 0.1 \mathcal{L}_{\text{KoLeo}}$, which we explain below.

\begin{itemize}[leftmargin=0.7cm, itemsep=0.5pt, topsep=0.5pt]
    \item 
\textbf{DINO Loss.} To encourage invariance to temporal scale and local noise, we employ a multi-crop strategy combined with random jittering. Given a time series input $x \in \mathbb{R}^{T}$, we generate the following augmentations: (a) two global random crops that cover from $40\% $ to $100\%$ of the signal resized to $T=512$, (b) eight local random crops that represent smaller segments with a $10\% $ to $40\%$ coverage of the signal and that are further resized to $T_{local}=256$. Gaussian jitter noise is applied randomly to some crops. 
The DINO objective minimizes the cross-entropy between the Student's and Teacher's [CLS] token probability distributions obtained via a shared-architecture projection head applied to each network's output. The Student is exposed to all augmented views, while the Teacher processes only global views. Probabilities are derived using softmax normalization. To prevent collapse, the Teacher's output is regularized using the Sinkhorn-Knopp algorithm for centering and a temperature parameter for sharpening.



\item \textbf{iBOT Loss.} To learn dense local features, we employ the iBOT objective. We apply patch-level masking to the global views fed into the Student network, with variable ratio drawn from a uniform distribution $\mathcal{U}(0.1, 0.7)$ and a sample probability of $0.5$. For a masked view $\hat{x}$, the Student predicts the token distribution of the masked patches, while the Teacher views the unmasked original signal. The loss is calculated as the cross-entropy between the Teacher and Student projected patch distributions at masked indices.

\item \textbf{KoLeo Regularizer \citep{sablayrolles2019spreadingvectorssimilaritysearch}.} To encourage a uniform distribution of features in the batch and prevent collapse, we apply the Kozachenko-Leonenko (KoLeo) differential entropy estimator to the Student's global [CLS] tokens before the projection head. 
\end{itemize}

\begin{figure}[ht]
\centering
\includegraphics[width=0.70\linewidth]{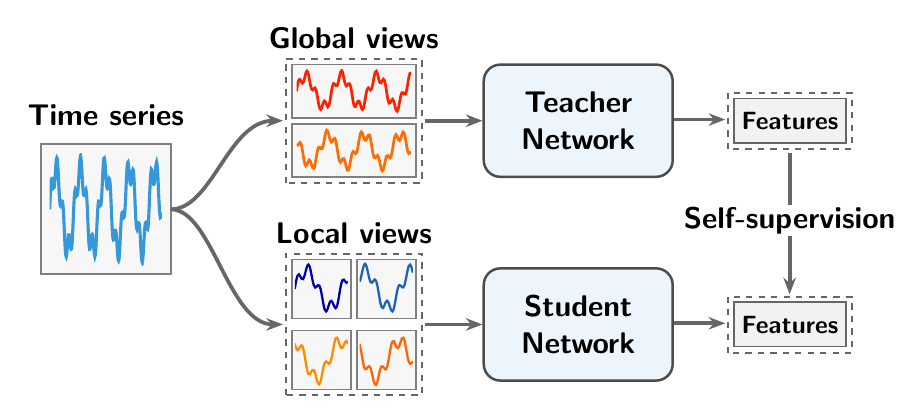}
\caption{\textbf{U}nlabeled \textbf{Ti}me-series \textbf{C}rop \textbf{A}ugmented framework. The self-supervised objective aims to match the features produced by the teacher with those produced by the student. Full details about pretraining can be found in Appendix \ref{app:pretraining}.}
\end{figure}

\section{Experimental Results}
In our experiments, we follow the Mantis evaluation protocol \citep{feofanov2025mantislightweightcalibratedfoundation}, considering two regimes: linear probing (frozen representations) and fine-tuning (end-to-end). 

\textbf{Baselines. } We compare UTICA, for which we use the Teacher network as the final model, against the following methods:
Mantis~\citep{feofanov2025mantislightweightcalibratedfoundation}, a 8M-parameter model trained with a contrastive loss;
Moment~\citep{goswami2024momentfamilyopentimeseries}, a 385M-parameter T5-based masked auto-encoder;
NuTime~\citep{lin2024nutimenumericallymultiscaledembedding}, a 2M-parameter Transformer encoder trained with self-distillation and 
GPT4TS \citep{zhou2023fitsallpowergeneraltime} a 80M-parameter partially fine-tuned GPT2.

\textbf{Datasets. }
We evaluate on the UCR archive~\citep{dau2019ucrtimeseriesarchive}, consisting of 128 univariate time series datasets, and the UEA archive~\citep{bagnall2018ueamultivariatetimeseries}, from which we use 21 multivariate datasets.

\begin{figure}[ht]
\centering
\includegraphics[width=0.8\textwidth]{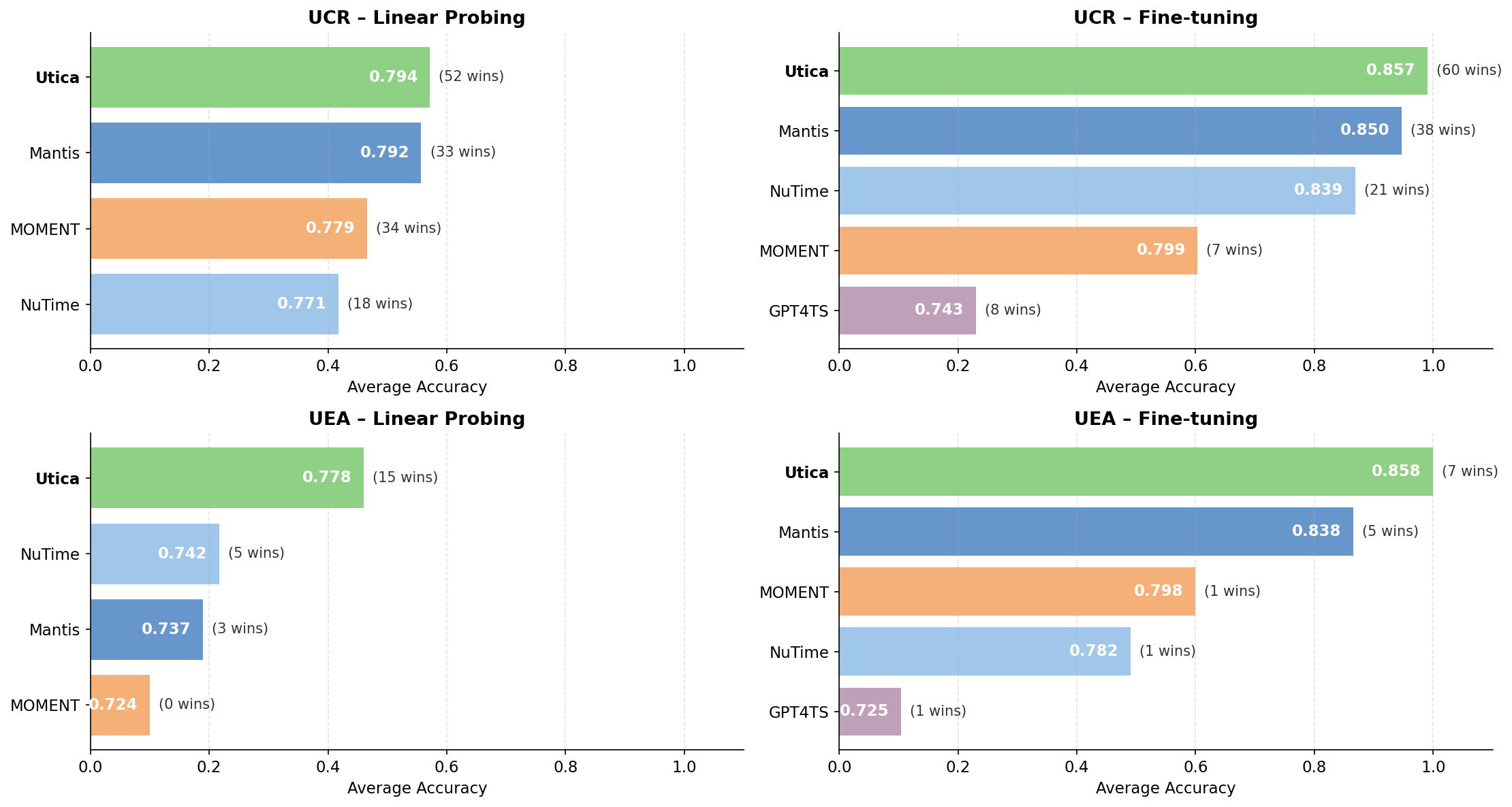}
\caption{Average accuracy comparison across linear probing and fine-tuning on both UCR and UEA benchmarks. Number of wins shown in parentheses.}
\label{fig:results}
\end{figure}


\textbf{Results. } Figure~\ref{fig:results} displays the obtained results on UCR and UEA when models are used in frozen and fine-tuned way (see the complete results in Appendix ~\ref{app:results}). Utica consistently outperforms all baselines across benchmarks and evaluation regimes. On UCR linear probing, Utica achieves $0.794$ average accuracy with $52/128$ wins, versus Mantis ($0.792$, $33$ wins) and MOMENT ($0.779$, $34$ wins). Under fine-tuning: Utica reaches $0.857$ with $60$ wins versus Mantis ($0.850$, $38$ wins). On UEA, Utica achieves the best average rank in both settings ($1.60$ linear probing, $1.50$ fine-tuning). 


\begin{table}[ht]
\caption{Ablation on the loss type used for pretraining.}
\label{tab:ablation-study}
\centering
\begin{tabular}{l|llll}
\toprule
Loss Type & data2vec & iBOT+KoLeo & DINO+KoLeo & UTICA \\ 
\midrule
Accuracy  & 0.7802   & 0.735      & 0.747      & \textbf{0.794}                                                             \\
\bottomrule
\end{tabular}
\end{table}

\textbf{Ablation Study. } In this section, we analyze our pretraining loss function more thoroughly. First, we study the contribution of the masking and scale-invariant loss components by pretraining the foundation model with iBOT+KoLeo and DINO+KoLeo, respectively. In Table \ref{tab:ablation-study}, we display the linear probing performance on UCR dataset. The experimental results reveal that local masked prediction (iBOT) and global multi-crop alignment (DINO) provide complementary supervision signals: individually, they have significantly lower performances (0.735 and 0.747, respectively) compared to their combination (0.794). 
In addition, we compare the proposed loss with data2vec \citep{baevski2022data2vecgeneralframeworkselfsupervised}, another self-distillation approach adapted for time series data \citep{pieper2023selfdistilledrepresentationlearningtime}. Under the same linear probing protocol, we can see that UTICA outperforms data2vec by $1.38\%$ on the UCR benchmark.



\section{Conclusion}

Our results demonstrate that an adaptation of  DINOv2-style self-distillation, previously successful in computer vision, transfers effectively to time series foundation models. The consistent results on both UCR and UEA benchmarks suggest that self-distillation pretraining is a promising direction for time series classification foundation models. Future work includes exploring alternative backbone architectures and scaling the model parameters.
\bibliography{references.bib}


\appendix

\newpage

\begin{center}
    {\LARGE \bfseries Appendix}
\end{center}
\vspace{1em}

\startcontents[appendix]

\renewcommand*\contentsname{\Large Table of Contents}

\printcontents[appendix]{}{1}{\setcounter{tocdepth}{2}}
\newpage

\section{Synthetic Data Generation}
\label{app:data}
We generate synthetic time series using a causal DAG. Root node signals are sampled from Gaussian Processes:
\[
x_v(t) \sim \mathcal{GP}(\mu_v(t), k_v(t,t')),
\]
where $\mu_v(t)$ is a non-stationary mean function and $k_v$ is a randomly composed covariance kernel.  

Non-root nodes combine their parent signals via a weighted sum followed by a nonlinear transformation:
\[
x_v(t) = f_v\!\left(\sum_{u \in \mathrm{pa}(v)} w_{uv} x_u(t) + b_v\right),
\]
with weights $w_{uv}$ and biases $b_v$ sampled from $\mathcal{N}(0,1)$ and $f_v$ a randomly sampled nonlinearity.

\begin{figure}[h!]
\centering
\includegraphics[width=0.6\linewidth]{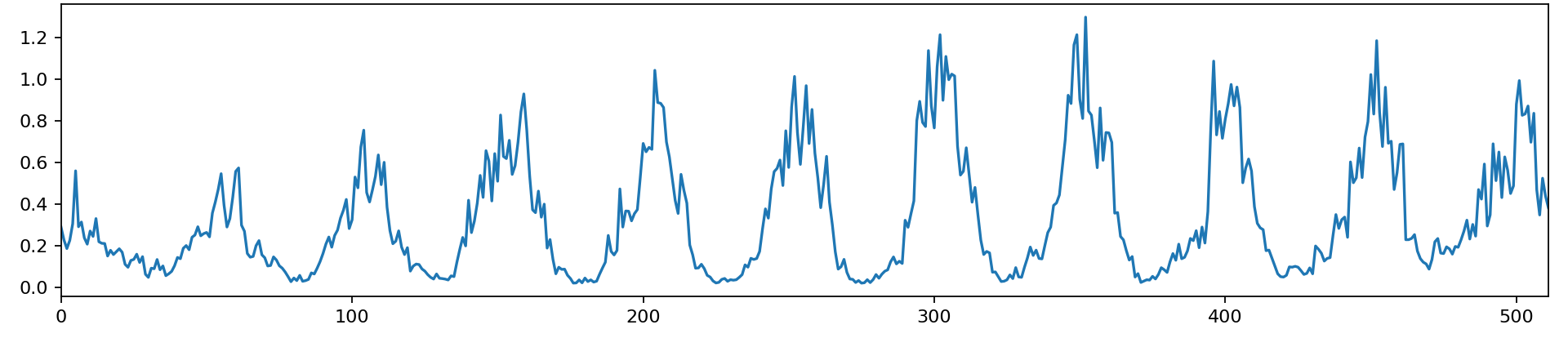}
\caption{Example of synthetic time series from the DAG-based generator.}
\label{fig:synthetic}
\end{figure}

A subset of node signals is observed to form the input time series used for pretraining.  
Figure~\ref{fig:synthetic} shows an example of the resulting synthetic sequences.

\section{Self-Supervised Pretraining Details}
\label{app:pretraining}

This appendix provides a comprehensive description of the self-supervised
pretraining procedure used to train the Mantis-8M time-series foundation model.
\subsection{Model Architecture}
\label{app:architecture}

Mantis-8M is a Transformer encoder operating on univariate time
series of length multiple of $32$.  The architecture consists of two stages:

\paragraph{Token Generator Unit.}
Each input series $\mathbf{x}\in\mathbb{R}^{T}$ is split into $P=32$
non-overlapping patches of window size $w = T/P = 16$.  Two convolutional
branches extract features from (i)~the z-score-normalised series and (ii)~its
first-order finite difference $\Delta\mathbf{x}$, both padded to preserve
length.  Each branch uses a 1-D convolution and output channels equal to
the hidden dimension $d=256$, followed by layer normalisation.  The branch
outputs are averaged within each patch to yield $P$ patch embeddings of
dimension~$d$.

In parallel, per-patch mean and standard deviation are encoded by
\emph{Multi-Scaled Scalar Encoders} with nine log-spaced scales
$\{10^{-4},\ldots,10^{4}\}$, each producing embeddings of dimension
$d_{\text{scalar}}=32$ (tolerance $\epsilon=1.1$).  The concatenated
convolutional and scalar embeddings are projected to $d=256$ by a linear
encoder.

\paragraph{Transformer Encoder Unit.}
The $P$ patch tokens are prepended with a learnable \texttt{[CLS]} token and
summed with sinusoidal positional encodings.  The
sequence is processed by a Transformer with the following hyper-parameters:

\begin{table}[ht]
\centering
\caption{Transformer hyper-parameters of the Transformer Encoder Unit.}
\label{tab:vit_params}
\begin{tabular}{lc}
\toprule
\textbf{Parameter} & \textbf{Value} \\
\midrule
Depth (number of layers) & 6 \\
Attention heads & 8 \\
Key/query/value dimension per head & 128 \\
MLP hidden dimension & 512 \\
Hidden dimension $d$ & 256 \\
Dropout & 0.1 \\
\bottomrule
\end{tabular}
\end{table}

\noindent
A final layer normalisation is applied, and the output is split into the
\texttt{[CLS]} token embedding $\mathbf{z}_{\texttt{cls}}\in\mathbb{R}^{d}$
and the patch token embeddings
$\mathbf{Z}_{\text{patch}}\in\mathbb{R}^{P\times d}$.

A learnable \texttt{[MASK]} token $\mathbf{m}\in\mathbb{R}^{d}$ (initialised
to zero) is used to replace masked patch tokens before they enter the
Transformer (see Section~\ref{app:masking}).

\subsection{Student--Teacher Framework}
\label{app:student_teacher}

We adopt the exponential moving average (EMA) student--teacher paradigm.  The
student and teacher share the same architecture (backbone + projection heads)
but differ in how they are updated:

\begin{itemize}
    \item \textbf{Student:} updated via gradient descent.
    \item \textbf{Teacher:} updated via EMA of the student parameters,
          $\boldsymbol{\theta}_t \leftarrow m\,\boldsymbol{\theta}_t
          + (1-m)\,\boldsymbol{\theta}_s$, where $m$ follows a cosine
          schedule from $m_0=0.992$ to $m_f=1.0$ over the course of training.
          The teacher is kept in evaluation mode with no gradient computation.
\end{itemize}

\subsection{Projection Heads}
\label{app:heads}

Both the DINO (CLS-token) and iBOT (patch-token) objectives use separate
head projectors with identical architectures:

\begin{table}[ht]
\centering
\caption{Projection head hyper-parameters (shared by DINO and iBOT heads).}
\label{tab:head_params}
\begin{tabular}{lc}
\toprule
\textbf{Parameter} & \textbf{Value} \\
\midrule
Input dimension & 256 \\
Hidden dimension & 2\,048 \\
Bottleneck dimension & 256 \\
Output dimension (prototypes) $K$ & 65\,536 \\
Number of MLP layers & 3 \\
Activation & GELU \\
Weight normalisation on last layer & $\ell_2$-normalisation before linear \\
\bottomrule
\end{tabular}
\end{table}

\noindent
The MLP maps the input to the bottleneck dimension, applies $\ell_2$
normalisation, and a final linear layer (without bias) projects to $K=65{,}536$
prototypes.  Weights are initialised with truncated normals ($\sigma=0.02$).

\subsection{Data Augmentation}
\label{app:augmentation}

Each univariate time series $\mathbf{x}\in\mathbb{R}^{T}$ undergoes
multi-crop augmentation:

\paragraph{Global crops.}
Two global crop are generated. A random
contiguous sub-sequence with  a fraction
$r \sim \mathcal{U}(0.4, 1.0)$ of the original length, then linearly
risized to $T_g=512$ via interpolation.  Gaussian jitter with
$\sigma_{\text{jitter}}=0.2\times\operatorname{std}(\mathbf{x})$ is applied to one of the global crops.

\paragraph{Local crops.}
Eight local crops are generated.  Each takes a contiguous sub-sequence with a 
fraction $r \sim \mathcal{U}(0.1, 0.4)$ and is resized to $T_\ell=256$ via interpolation.  Jitter is
applied with probability $p=0.5$.

\begin{figure}[h]
    \centering
    \begin{minipage}[t]{0.48\linewidth}
        \centering
        \includegraphics[width=\linewidth]{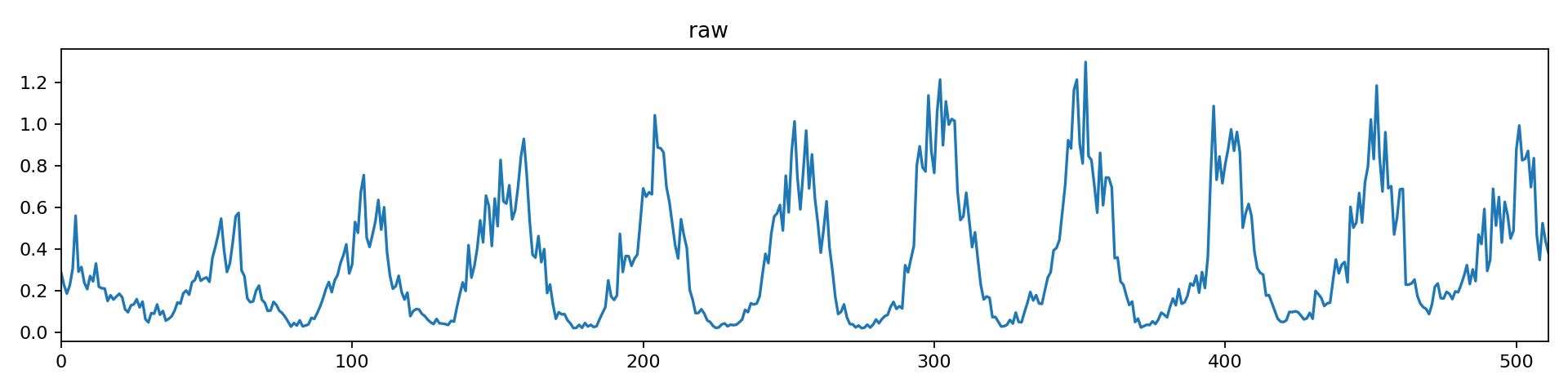}
        \caption{Original time series.}
        \label{fig:original}
    \end{minipage}\hfill
    \begin{minipage}[h]{0.48\linewidth}
        \centering
        \includegraphics[width=\linewidth]{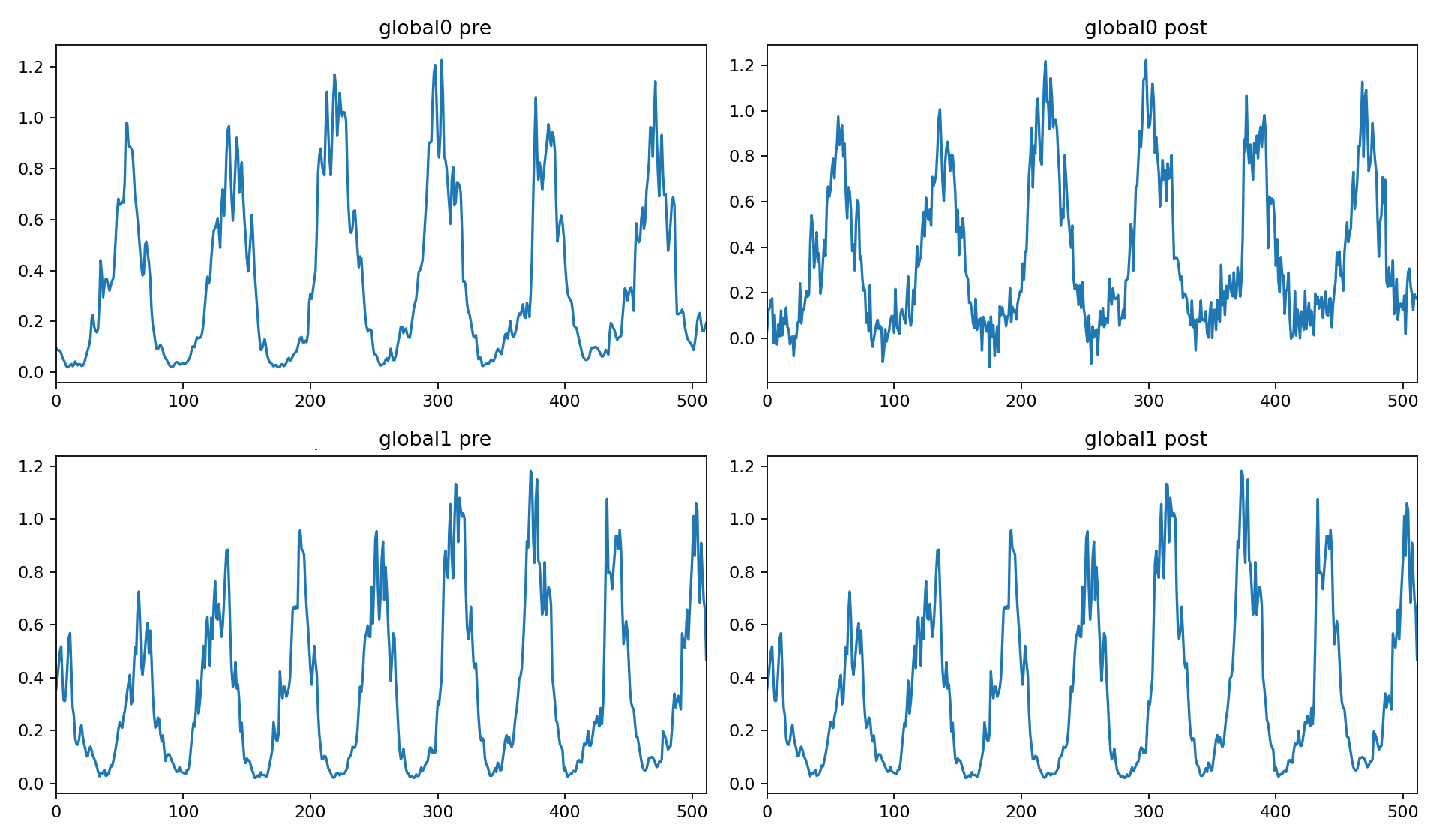}
        \caption{Two global views before and after eventual jittering.}
        \label{fig:global}
    \end{minipage}

    \vspace{0.5em}

    \includegraphics[width=0.6\linewidth]{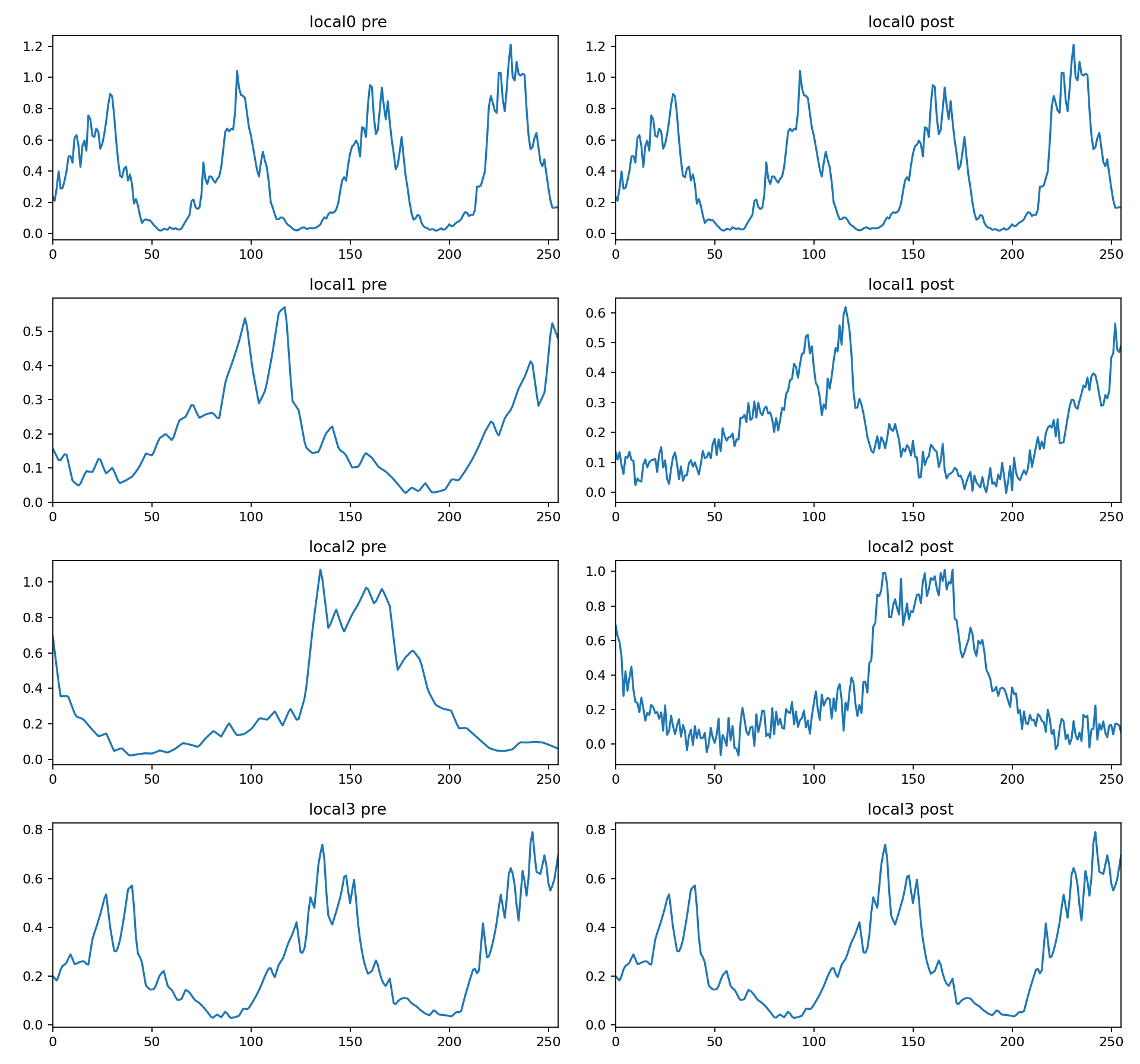}
    \caption{Four local views before and after eventual jittering.}
    \label{fig:local}
\end{figure}

\subsection{Masking Strategy}
\label{app:masking}

Patch-level masking is applied to the global crops seen by the student to
enable the iBOT patch-reconstruction objective.  Given $P=32$ patches:

\begin{enumerate}
    \item Each sample in the batch is selected for masking with probability
          $p_{\text{mask}}=0.5$.
    \item For selected samples, the mask ratio is drawn from a linearly spaced
          schedule between $r_{\min}=0.1$ and $r_{\max}=0.7$, yielding between
          $\lfloor 0.1 \times 32 \rfloor = 3$ and
          $\lfloor 0.7 \times 32 \rfloor = 22$ masked patches.
    \item Masked patch positions are chosen uniformly at random (without
          replacement).
    \item Unselected samples receive an all-zero mask (no patches masked).
\end{enumerate}

\noindent
Masked patch tokens in the student input are replaced by the learnable
\texttt{[MASK]} token before entering the Transformer.  The teacher always sees
unmasked inputs.

\begin{figure}[h]
    \centering
    \includegraphics[width=0.5\linewidth]{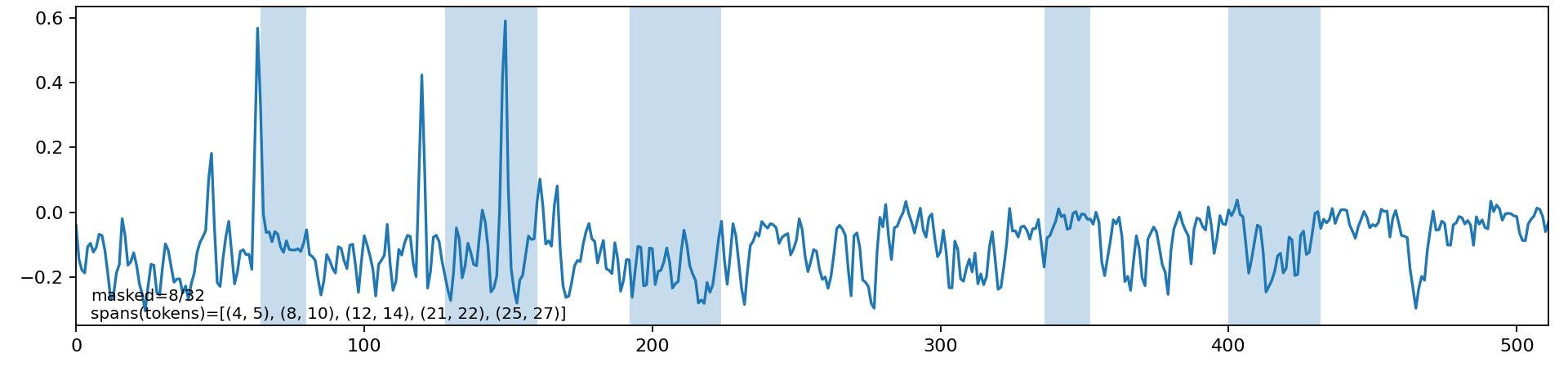}
    \caption{Masked patches example}
    \label{fig:masked_patches}
\end{figure}

\subsection{Loss Functions}
\label{app:losses}

The total loss is a weighted combination of three terms:

\begin{equation*}
    \mathcal{L} = \lambda_{\text{DINO}}\,\mathcal{L}_{\text{DINO}}
                 + \lambda_{\text{iBOT}}\,\mathcal{L}_{\text{iBOT}}
                 + \lambda_{\text{KoLeo}}\,\mathcal{L}_{\text{KoLeo}},
\end{equation*}

\noindent with $\lambda_{\text{DINO}}=1.0$, $\lambda_{\text{iBOT}}=1.0$, and
$\lambda_{\text{KoLeo}}=0.1$.

\paragraph{DINO loss.}
The DINO loss is the cross-entropy between teacher and student probability
distributions over $K=65{,}536$ prototypes, computed on \texttt{[CLS]} tokens.

\subparagraph{Teacher targets via Sinkhorn--Knopp.}
Let $\mathbf{z}_t^{(i)} \in \mathbb{R}^K$ denote the teacher head output for
sample~$i$ in teacher crop~$t$.  We form the soft-assignment matrix
$\mathbf{Q} \in \mathbb{R}^{K \times B_{\text{eff}}}$ (prototypes $\times$
samples) as:
\begin{equation*}
    Q_{k,i} = \frac{\exp\bigl(z_{t,k}^{(i)} / \tau_t\bigr)}
              {\sum_{k',i'}\exp\bigl(z_{t,k'}^{(i')} / \tau_t\bigr)},
\end{equation*}
where the denominator includes a distributed all-reduce so the matrix sums to~1
across all GPUs.  Three iterations of Sinkhorn--Knopp alternating
row-normalisation (each row sums to $1/K$) and column-normalisation (each column
sums to $1/B_{\text{eff}}$) are then applied, where
$B_{\text{eff}} = B \times \text{world\_size}$ is the effective batch size.
After convergence the columns are rescaled to sum to~1, yielding the teacher
probability vectors
$\mathbf{p}_t^{(i)} = \mathbf{Q}_{:,i} \in \Delta^{K-1}$.

The teacher temperature $\tau_t$ follows a linear warmup from
$\tau_{t,0}=0.04$ to $\tau_t=0.07$ over the first 2.5 epochs.

\subparagraph{Student probabilities.}
The student head output $\mathbf{z}_s^{(i)}$ for student crop~$s$ is converted
to log-probabilities:
\begin{equation*}
    \mathbf{q}_s^{(i)} = \log\operatorname{softmax}\!\bigl(\mathbf{z}_s^{(i)}
    / \tau_s\bigr), \qquad \tau_s = 0.1.
\end{equation*}

\subparagraph{Cross-entropy computation.}
The per-pair cross-entropy between student crop~$s$ and teacher crop~$t$ is:
\begin{equation*}
    H(s,t) = -\frac{1}{B}\sum_{i=1}^{B}\sum_{k=1}^{K}
    p_{t,k}^{(i)}\;\log q_{s,k}^{(i)}.
\end{equation*}

The total DINO loss distinguishes two terms, following the DINOv2 scaling
convention.  Let $S_g$ and $S_\ell$ denote the number of global and local
student crops, and $T_g$ the number of teacher (global) crops.

\textit{Global--global term}: a global student crop is not compared
with the same-index teacher crop.  The raw loss matrix
$L_{s,t} = H(s,t)$ for $s,t \in \{1,\ldots,S_g\}$ has its diagonal set to
zero before summation.  The effective number of terms is
$n_g = S_g \cdot T_g - \min(S_g, T_g)$:
\begin{equation*}
    \mathcal{L}_{\text{DINO}}^{\text{global}} =
    \frac{1}{B\,n_g}\sum_{\substack{s,t=1\\s\neq t}}^{S_g,\,T_g}
    \Bigl(-\sum_{i,k} p_{t,k}^{(i)}\,\log q_{s,k}^{(i)}\Bigr).
\end{equation*}

\textit{Local--global term}: all local student crops are compared with all
teacher global crops.  The number of terms is
$n_\ell = S_\ell \cdot T_g$:
\begin{equation*}
    \mathcal{L}_{\text{DINO}}^{\text{local}} =
    \frac{1}{B\,n_\ell}\sum_{s=1}^{S_\ell}\sum_{t=1}^{T_g}
    \Bigl(-\sum_{i,k} p_{t,k}^{(i)}\,\log q_{s,k}^{(i)}\Bigr).
\end{equation*}

The two terms are combined with scaling factors that reflect the fraction of
cross-crop comparisons each contributes:
\begin{equation*}
    \alpha_g = \frac{n_g}{n_g + n_\ell}, \qquad
    \alpha_\ell = \frac{n_\ell}{n_g + n_\ell}, \qquad
    \mathcal{L}_{\text{DINO}} =
    \alpha_g \,\mathcal{L}_{\text{DINO}}^{\text{global}}
    + \alpha_\ell \,\mathcal{L}_{\text{DINO}}^{\text{local}}.
\end{equation*}

\paragraph{iBOT patch loss.}
The iBOT objective operates on individual patch tokens rather than the
\texttt{[CLS]} token.  It is computed only over patches that are masked for the
student.

\subparagraph{Teacher targets via Sinkhorn--Knopp (patch-level).}
Let $M = \sum_{i=1}^{B}\sum_{j=1}^{P} \mathbb{1}[\text{mask}_{i,j}=1]$ be the
total number of masked patches across all samples and GPUs.  The teacher backbone
processes unmasked global crops; the patch tokens at masked positions are
extracted and passed through the iBOT head to produce
$\hat{\mathbf{z}}_m \in \mathbb{R}^K$ for each masked patch $m=1,\ldots,M$.

Sinkhorn--Knopp is applied identically to the DINO case:
\begin{equation*}
    Q_{k,m}^{\text{patch}} \propto \exp\!\bigl(\hat{z}_{m,k}/\tau_t\bigr),
\end{equation*}
with 3 iterations of alternating row- and column-normalisation using
distributed all-reduces.  This yields teacher patch probabilities
$\hat{\mathbf{p}}_m \in \Delta^{K-1}$.

\subparagraph{Student predictions.}
The student receives the global crop with masked patches replaced by the
learnable \texttt{[MASK]} token.  The output patch tokens at masked positions
are extracted and passed through the student iBOT head, yielding logits
$\hat{\mathbf{s}}_m \in \mathbb{R}^K$.

\subparagraph{Weighted cross-entropy.}
The per-patch cross-entropy is:
\begin{equation*}
    \ell_m = -\sum_{k=1}^{K} \hat{p}_{m,k}\;
    \log\operatorname{softmax}_k\!\bigl(\hat{\mathbf{s}}_m / \tau_s\bigr),
    \qquad \tau_s = 0.1.
\end{equation*}

To ensure that each \emph{sample} contributes equally regardless of how many of
its patches are masked, each patch loss is weighted by the inverse of the number
of masked patches in its parent sample.  Concretely, let $M_i$ denote the
number of masked patches in sample~$i$, and let $\sigma(m) = i$ map each masked
patch index to its parent sample.  The weight is:
\begin{equation*}
    w_m = \frac{1}{\max(M_{\sigma(m)},\;1)}.
\end{equation*}

The final iBOT loss is normalised by the number
of global-crop samples~$B$:
\begin{equation*}
    \mathcal{L}_{\text{iBOT}} = -\frac{1}{B}\sum_{m=1}^{M} w_m \,\ell_m.
\end{equation*}

\paragraph{KoLeo loss.}
The Kozachenko--Leonenko entropy estimator is applied to the $\ell_2$-normalised
pre-head \texttt{[CLS]} tokens of the student's global crops:

\begin{equation*}
    \mathcal{L}_{\text{KoLeo}} = -\frac{1}{B}\sum_{i=1}^{B}
    \log\bigl(\min_{j\neq i}\|\hat{\mathbf{z}}_i - \hat{\mathbf{z}}_j\|_2
    + \epsilon\bigr),
\end{equation*}

\noindent where $\hat{\mathbf{z}}$ denotes the $\ell_2$-normalised
representation and $\epsilon=10^{-8}$.

\subsection{Optimisation}
\label{app:optimisation}

\begin{table}[ht]
\centering
\caption{Optimisation hyper-parameters.}
\label{tab:optim}
\begin{tabular}{lc}
\toprule
\textbf{Parameter} & \textbf{Value} \\
\midrule
Optimiser & AdamW \\
$(\beta_1,\;\beta_2)$ & $(0.9,\;0.999)$ \\
Base learning rate & $1\times10^{-3}$ \\
Minimum learning rate & $1\times10^{-7}$ \\
Learning rate schedule & Cosine decay \\
Warmup epochs & 0.7 \\
Weight decay (start $\to$ end) & $0.04 \to 0.4$ (cosine schedule) \\
Gradient clipping  & 3.0 \\
Layer-wise learning rate decay & 0.9 \\
Patch embed learning rate multiplier & 0.2 \\
Freeze last layer epochs & 0.07 \\
\bottomrule
\end{tabular}
\end{table}

\noindent
The learning rate and weight decay follow cosine schedules.  Layer-wise
learning rate decay is applied to the Transformer Encoder layers: layer $l$ (out of $L=6$)
receives a learning rate multiplied by $0.9^{L+1-l}$.  Parameters in the token
generator unit additionally receive a $0.2\times$ multiplier.  Bias, norm, and
gamma parameters have their weight decay set to zero.

The final linear projection in each projection head (the layer mapping from the
bottleneck to the $K=65{,}536$ prototypes) follows a separate learning rate
schedule.  This schedule is identical to the main cosine learning rate schedule,
except that its values are set to zero for the first iterations.  During this
period the prototype weights receive no gradient updates, while all other
parameters train normally.  This stabilises early training.

\section{Downstream Evaluation}
\subsection{Linear Probing}
\sloppy
Linear classifier trained on frozen backbone embeddings.  
\begin{itemize}
    \item Optimizer: AdamW
    \item Epochs: 100
    \item Learning rate: Fixed to 1e-3
\end{itemize}

\subsection{Fine-Tuning}
Following \citep{feofanov2025mantislightweightcalibratedfoundation} protocol for full model fine-tuning:
\begin{itemize}
    \item Optimizer: AdamW
    \item Scheduler: Cosine
    \item Weight decay: 0.05
    \item Epochs: 100
    \item Learning rate: grid search on $[1e-4, 2e-4, 1e-3]$, used $20\%$ of training data as validation 
\end{itemize}

\section{Experimental Results}\label{app:results}
\subsection{Per-Dataset Performance}

\begin{table}
    \centering
    \caption{Comparison under linear probing regime on UEA.
The best epoch performance is averaged over 3 random seeds and reported with the standard deviation.
Bold indicates the best result per row. NuTime, Moment and Mantis results are from \citep{feofanov2025mantislightweightcalibratedfoundation}. We exclude datasets whose memory requirements exceed our available GPU memory (24\,GB).}
    \label{tab:performance_comparison}
    \resizebox{\textwidth}{!}{%
    \begin{tabular}{lcccc}
        \toprule
        \textbf{Dataset} & \textbf{NuTime} & \textbf{Moment} & \textbf{Mantis} & \textbf{Utica} \\
        \midrule
        ArticularyWordRecognition & \textbf{0.9933} $_{\pm 0.0000}$ & 0.9800 $_{\pm 0.0000}$ & 0.9930 $_{\pm 0.0030}$ & 0.9900  $_{\pm 0.0027}$ \\
        BasicMotions & \textbf{1.0000} $_{\pm 0.0000}$ & 0.9920 $_{\pm 0.0140}$ & \textbf{1.0000} $_{\pm 0.0000}$ & \textbf{1.0000} $_{\pm 0.0000}$ \\
        CharacterTrajectories & 0.9649 $_{\pm 0.0038}$ & 0.9700 $_{\pm 0.0020}$ & 0.9400 $_{\pm 0.0010}$ & \textbf{0.9847} $_{\pm 0.0006}$\\
        Cricket & 0.9907 $_{\pm 0.0080}$ & 0.9770 $_{\pm 0.0080}$ & \textbf{1.0000} $_{\pm 0.0000}$ & 0.9907 $_{\pm 0.0066}$ \\
        ERing & \textbf{0.9716} $_{\pm 0.0057}$ & 0.9690 $_{\pm 0.0080}$ & 0.9410 $_{\pm 0.0100}$ & 0.9605 $_{\pm 0.0017}$ \\
        Eigenworms & 0.7786 $_{\pm 0.0202}$ & 0.7350 $_{\pm 0.0160}$ & 0.7530 $_{\pm 0.0160}$ & \textbf{0.7837} $_{\pm 0.0190}$ \\
        Epilepsy & \textbf{1.0000} $_{\pm 0.0000}$ & 0.9880 $_{\pm 0.0040}$ & 0.9950 $_{\pm 0.0040}$ & 0.9855  $_{\pm 0.0000}$ \\
        EthanolConcentration & \textbf{0.3942} $_{\pm 0.0158}$ & 0.2780 $_{\pm 0.0080}$ & 0.2700 $_{\pm 0.0100}$ & 0.3561  $_{\pm 0.0100}$\\
        FingerMovements & 0.5133 $_{\pm 0.0153}$ & 0.4970 $_{\pm 0.0230}$ & 0.5400 $_{\pm 0.0350}$ & \textbf{0.5733} $_{\pm 0.0094}$  \\
        HandMovementDirection & 0.3108 $_{\pm 0.0234}$ & 0.3150 $_{\pm 0.0080}$ & 0.2120 $_{\pm 0.0410}$ & \textbf{0.3514} $_{\pm 0.0441}$\\
        Handwriting & 0.2035 $_{\pm 0.0082}$ & 0.2360 $_{\pm 0.0050}$ & 0.3390 $_{\pm 0.0100}$ & \textbf{0.4227}  $_{\pm 0.0095}$ \\
        JapaneseVowels & 0.9405 $_{\pm 0.0072}$ & 0.8800 $_{\pm 0.0090}$ & 0.9630 $_{\pm 0.0060}$ & \textbf{0.9784} $_{\pm 0.0022}$\\
        LSST & 0.5668 $_{\pm 0.0068}$ & 0.6210 $_{\pm 0.0010}$ & 0.6050 $_{\pm 0.0020}$ & \textbf{0.6919 $_{\pm 0.0019}$} \\
        Libras & 0.8852 $_{\pm 0.0064}$ & 0.8500 $_{\pm 0.0200}$ & 0.8980 $_{\pm 0.0080}$ & \textbf{0.9000} $_{\pm 0.0046}$\\
        NATOPS & 0.8426 $_{\pm 0.0140}$ & 0.8220 $_{\pm 0.0240}$ & 0.9070 $_{\pm 0.0130}$ & \textbf{0.9185} $_{\pm 0.0026}$\\
        PhonemeSpectra & 0.2588 $_{\pm 0.0065}$ & 0.2150 $_{\pm 0.0010}$ & 0.2730 $_{\pm 0.0080}$ & \textbf{0.2982} $_{\pm 0.0007}$\\
        RacketSports & 0.9101 $_{\pm 0.0100}$ & 0.8200 $_{\pm 0.0100}$ & \textbf{0.9230} $_{\pm 0.0040}$ & 0.8838  $_{\pm 0.0135}$ \\
        SelfRegulationSCP1 & 0.7702 $_{\pm 0.0120}$ & 0.7540 $_{\pm 0.0120}$ & 0.8040 $_{\pm 0.0110}$ & \textbf{0.8612} $_{\pm 0.0126}$ \\
        SelfRegulationSCP2 & 0.5000 $_{\pm 0.0242}$ & 0.4960 $_{\pm 0.0200}$ & 0.4760 $_{\pm 0.0450}$ & \textbf{0.5315} $_{\pm 0.0052}$ \\
        SpokenArabicDigits & 0.8968 $_{\pm 0.0032}$ & 0.9350 $_{\pm 0.0020}$ & 0.8390 $_{\pm 0.0050}$ & \textbf{0.9891}   $_{\pm 0.0004}$\\
        UWaveGestureLibrary & 0.8792 $_{\pm 0.0018}$ & 0.8730 $_{\pm 0.0110}$ & 0.8140 $_{\pm 0.0100}$ & \textbf{0.8802}  $_{\pm 0.0103}$\\
        \midrule
        \textbf{Total Wins} & 5 & 0 & 3 & \textbf{15} \\
        \textbf{Average Rank} & 2.45 &3.29 & 2.67 & \textbf{1.60} \\
        \textbf{Average Accuracy} & 0.7415 & 0.7240 & 0.7374 & \textbf{0.7777} \\
        \bottomrule
    \end{tabular}%
    }
\end{table}

\begin{table}
\centering
    \caption{Comparison under the fine tuning regime on UEA.
The best epoch performance is averaged over 3 random seeds and reported with the standard deviation.
Bold indicates the best result per row. GPT4S, NuTime, Moment and Mantis results are from \citep{feofanov2025mantislightweightcalibratedfoundation}. We exclude datasets whose memory requirements exceed our available GPU memory (24\,GB).}
\resizebox{\textwidth}{!}{%
\begin{tabular}{lccccc}
\toprule
Dataset & GPT4TS & NuTime & Moment & Mantis & Utica \\
\midrule
ArticularyWordRecognition & $0.9633_{\pm 0.0058}$ & $0.9833_{\pm 0.0033}$ & $0.9733_{\pm 0.0033}$ & \textbf{0.9944$_{\pm 0.0019}$} & $0.9900_{\pm 0.0033}$ \\
BasicMotions & $0.8833_{\pm 0.0144}$ & \textbf{1.0000$_{\pm 0.0000}$} & \textbf{1.0000$_{\pm 0.0000}$} & \textbf{1.0000$_{\pm 0.0000}$} & \textbf{1.0000$_{\pm 0.0000}$} \\
CharacterTrajectories & $0.9731_{\pm 0.0056}$ & $0.9800_{\pm 0.0036}$ & $0.9791_{\pm 0.0028}$ & $0.9893_{\pm 0.0008}$ & \textbf{0.9937$_{\pm 0.0007}$} \\
Cricket & $0.9213_{\pm 0.0080}$ & $0.9907_{\pm 0.0080}$ & $0.9491_{\pm 0.0160}$ & $0.9954_{\pm 0.0080}$ & \textbf{1.0000$_{\pm 0.0000}$} \\
ERing & $0.9074_{\pm 0.0098}$ & $0.9506_{\pm 0.0077}$ & $0.9358_{\pm 0.0077}$ & $0.9827_{\pm 0.0043}$ & \textbf{0.9864$_{\pm 0.0021}$} \\
Epilepsy & $0.8092_{\pm 0.0233}$ & $0.9952_{\pm 0.0042}$ & $0.9976_{\pm 0.0042}$ & \textbf{1.0000$_{\pm 0.0000}$} & $0.9952_{\pm 0.0042}$ \\
HandMovementDirection & \textbf{0.4685$_{\pm 0.0680}$} & $0.3153_{\pm 0.0281}$ & $0.3108_{\pm 0.0358}$ & $0.3108_{\pm 0.0702}$ & $0.4189_{\pm 0.0135}$ \\
Handwriting & $0.2922_{\pm 0.0049}$ & $0.2545_{\pm 0.0118}$ & $0.3635_{\pm 0.0101}$ & \textbf{0.4631$_{\pm 0.0147}$} & $0.4451_{\pm 0.0226}$ \\
LSST & $0.1080_{\pm 0.0611}$ & $0.3335_{\pm 0.0282}$ & $0.5827_{\pm 0.0162}$ & $0.6035_{\pm 0.0254}$ & \textbf{0.7091$_{\pm 0.0075}$} \\
Libras & $0.7333_{\pm 0.0096}$ & $0.8407_{\pm 0.0064}$ & $0.7852_{\pm 0.0064}$ & $0.8722_{\pm 0.0056}$ & \textbf{0.9019$_{\pm 0.0032}$} \\
RacketSports & $0.7961_{\pm 0.0174}$ & $0.8794_{\pm 0.0137}$ & $0.7895_{\pm 0.0132}$ & \textbf{0.9189$_{\pm 0.0100}$} & \textbf{0.9189$_{\pm 0.0082}$} \\
UWaveGestureLibrary & $0.8396_{\pm 0.0065}$ & $0.8646_{\pm 0.0188}$ & $0.9156_{\pm 0.0094}$ & $0.9281_{\pm 0.0125}$ & \textbf{0.9406$_{\pm 0.0125}$} \\
\midrule
Wins & 1 & 1 & 1 & 5 & \textbf{7} \\
Avg Rank & 4.50 & 3.17 & 3.42 & 1.75 & \textbf{1.50} \\
Avg Accuracy & 0.7246 & 0.7823 & 0.7985 & 0.8382 & \textbf{0.8583} \\
\bottomrule
\end{tabular}
}
\end{table}


\begin{longtable}{lcccc}
\caption{Comparison under linear probing regime on UCR.
The best epoch performance is averaged over 3 random seeds and reported with the standard deviation.
Bold indicates the best result per row. NuTime, Moment and Mantis results are from \citep{feofanov2025mantislightweightcalibratedfoundation}.} \\
\toprule
\textbf{Dataset} & \textbf{NuTime} & \textbf{Moment} & \textbf{Mantis} & \textbf{Utica} \\
\midrule
\endfirsthead
\multicolumn{5}{c}{\tablename\ \thetable{} -- \textit{continued from previous page}} \\
\toprule
\textbf{Dataset} & \textbf{NuTime} & \textbf{Moment} & \textbf{Mantis} & \textbf{Utica} \\
\midrule
\endhead
\midrule
\multicolumn{5}{r}{\textit{Continued on next page}} \\
\endfoot
\bottomrule
\endlastfoot
ACSF1 & 0.7367$_{\pm 0.0351}$ & \textbf{0.75$_{\pm 0.0173}$} & 0.6133$_{\pm 0.0208}$ & 0.6200$_{\pm 0.0000}$ \\
Adiac & 0.7255$_{\pm 0.0115}$ & \textbf{0.7886$_{\pm 0.0074}$} & 0.7332$_{\pm 0.0039}$ & 0.5090$_{\pm 0.0142}$ \\
AllGestureWiimoteX & 0.661$_{\pm 0.003}$ & 0.6105$_{\pm 0.0136}$ & 0.6705$_{\pm 0.0044}$ & \textbf{0.6824$_{\pm 0.0068}$} \\
AllGestureWiimoteY & 0.6362$_{\pm 0.0079}$ & 0.6576$_{\pm 0.0079}$ & 0.6671$_{\pm 0.0057}$ & \textbf{0.7029$_{\pm 0.0071}$} \\
AllGestureWiimoteZ & 0.6024$_{\pm 0.0179}$ & 0.5767$_{\pm 0.0103}$ & \textbf{0.6695$_{\pm 0.0033}$} & 0.6448$_{\pm 0.0016}$ \\
ArrowHead & 0.76$_{\pm 0.0198}$ & \textbf{0.8076$_{\pm 0.0144}$} & 0.7105$_{\pm 0.0175}$ & 0.6514$_{\pm 0.0618}$ \\
BME & 0.8444$_{\pm 0.0168}$ & 0.9756$_{\pm 0.0139}$ & 0.9311$_{\pm 0.0038}$ & \textbf{0.9933$_{\pm 0.0000}$} \\
Beef & 0.6556$_{\pm 0.077}$ & \textbf{0.7444$_{\pm 0.0509}$} & 0.6556$_{\pm 0.0509}$ & 0.6111$_{\pm 0.0192}$ \\
BeetleFly & 0.8667$_{\pm 0.0289}$ & \textbf{0.95$_{\pm 0.0}$} & 0.85$_{\pm 0.05}$ & 0.8833$_{\pm 0.0577}$ \\
BirdChicken & 0.9667$_{\pm 0.0289}$ & 0.85$_{\pm 0.0}$ & \textbf{1.0$_{\pm 0.0}$} & 0.9000$_{\pm 0.0000}$ \\
CBF & 0.9744$_{\pm 0.0011}$ & 0.9411$_{\pm 0.0078}$ & 0.993$_{\pm 0.0013}$ & \textbf{0.9952$_{\pm 0.0006}$} \\
Car & 0.75$_{\pm 0.0}$ & \textbf{0.7944$_{\pm 0.0255}$} & 0.7722$_{\pm 0.0419}$ & 0.7611$_{\pm 0.0347}$ \\
Chinatown & 0.932$_{\pm 0.0017}$ & \textbf{0.9806$_{\pm 0.0034}$} & 0.8737$_{\pm 0.0168}$ & 0.9660$_{\pm 0.0034}$ \\
ChlorineConcentration & 0.6675$_{\pm 0.0035}$ & \textbf{0.6901$_{\pm 0.0042}$} & 0.6806$_{\pm 0.0019}$ & 0.5729$_{\pm 0.0021}$ \\
CinCECGTorso & 0.737$_{\pm 0.0152}$ & 0.6998$_{\pm 0.0118}$ & 0.6611$_{\pm 0.0036}$ & \textbf{0.7430$_{\pm 0.0123}$} \\
Coffee & 0.9167$_{\pm 0.0206}$ & 0.8929$_{\pm 0.0}$ & 0.9524$_{\pm 0.0206}$ & \textbf{0.9881$_{\pm 0.0206}$} \\
Computers & \textbf{0.78$_{\pm 0.004}$} & 0.6173$_{\pm 0.0162}$ & 0.7373$_{\pm 0.0092}$ & 0.7600$_{\pm 0.0040}$ \\
CricketX & 0.6701$_{\pm 0.0146}$ & 0.6795$_{\pm 0.0051}$ & 0.7368$_{\pm 0.0171}$ & \textbf{0.7436$_{\pm 0.0092}$} \\
CricketY & 0.6556$_{\pm 0.0053}$ & 0.6897$_{\pm 0.0077}$ & \textbf{0.7504$_{\pm 0.0065}$} & 0.7350$_{\pm 0.0015}$ \\
CricketZ & 0.6863$_{\pm 0.0171}$ & 0.7128$_{\pm 0.0044}$ & \textbf{0.7906$_{\pm 0.0039}$} & 0.7778$_{\pm 0.0121}$ \\
Crop & 0.6683$_{\pm 0.0026}$ & \textbf{0.7035$_{\pm 0.0026}$} & 0.6756$_{\pm 0.0018}$ & 0.6960$_{\pm 0.0011}$ \\
DiatomSizeReduction & 0.8322$_{\pm 0.0115}$ & \textbf{0.8867$_{\pm 0.0019}$} & 0.8845$_{\pm 0.0019}$ & 0.8322$_{\pm 0.0019}$ \\
DistalPhalanxOutlineAgeGroup & 0.7362$_{\pm 0.011}$ & 0.7506$_{\pm 0.011}$ & \textbf{0.789$_{\pm 0.015}$} & 0.7770$_{\pm 0.0000}$ \\
DistalPhalanxOutlineCorrect & 0.7742$_{\pm 0.0055}$ & \textbf{0.7886$_{\pm 0.0055}$} & 0.75$_{\pm 0.0126}$ & 0.7621$_{\pm 0.0021}$ \\
DistalPhalanxTW & 0.6763$_{\pm 0.0}$ & 0.6571$_{\pm 0.011}$ & 0.6859$_{\pm 0.011}$ & \textbf{0.6906$_{\pm 0.0000}$} \\
DodgerLoopDay & 0.5167$_{\pm 0.0361}$ & 0.4542$_{\pm 0.0144}$ & 0.55$_{\pm 0.0217}$ & \textbf{0.6167$_{\pm 0.0191}$} \\
DodgerLoopGame & 0.756$_{\pm 0.0084}$ & 0.8116$_{\pm 0.0126}$ & 0.7585$_{\pm 0.0221}$ & \textbf{0.8720$_{\pm 0.0042}$} \\
DodgerLoopWeekend & 0.9565$_{\pm 0.0072}$ & 0.9614$_{\pm 0.0111}$ & 0.9517$_{\pm 0.0084}$ & \textbf{0.9831$_{\pm 0.0042}$} \\
ECG200 & 0.8133$_{\pm 0.0153}$ & \textbf{0.8967$_{\pm 0.0153}$} & 0.82$_{\pm 0.01}$ & 0.8167$_{\pm 0.0153}$ \\
ECG5000 & 0.9313$_{\pm 0.0003}$ & \textbf{0.9384$_{\pm 0.0008}$} & 0.9211$_{\pm 0.001}$ & 0.9381$_{\pm 0.0011}$ \\
ECGFiveDays & 0.7801$_{\pm 0.017}$ & 0.8564$_{\pm 0.0223}$ & 0.909$_{\pm 0.0218}$ & \textbf{0.9311$_{\pm 0.0367}$} \\
EOGHorizontalSignal & 0.4346$_{\pm 0.0032}$ & 0.5571$_{\pm 0.0112}$ & \textbf{0.5875$_{\pm 0.0089}$} & 0.5737$_{\pm 0.0084}$ \\
EOGVerticalSignal & 0.2716$_{\pm 0.008}$ & 0.4595$_{\pm 0.0097}$ & 0.4751$_{\pm 0.0}$ & \textbf{0.4908$_{\pm 0.0112}$} \\
Earthquakes & 0.7458$_{\pm 0.0042}$ & 0.7458$_{\pm 0.0042}$ & 0.7482$_{\pm 0.0}$ & \textbf{0.7530$_{\pm 0.0083}$} \\
ElectricDevices & 0.7046$_{\pm 0.0014}$ & 0.7142$_{\pm 0.001}$ & \textbf{0.7226$_{\pm 0.0026}$} & 0.7107$_{\pm 0.0014}$ \\
EthanolLevel & 0.3407$_{\pm 0.0101}$ & \textbf{0.4227$_{\pm 0.0058}$} & 0.2993$_{\pm 0.011}$ & 0.3273$_{\pm 0.0046}$ \\
FaceAll & 0.6363$_{\pm 0.0053}$ & 0.7398$_{\pm 0.0074}$ & \textbf{0.7815$_{\pm 0.0074}$} & 0.7544$_{\pm 0.0045}$ \\
FaceFour & 0.7841$_{\pm 0.0521}$ & 0.7765$_{\pm 0.0174}$ & \textbf{0.9508$_{\pm 0.0066}$} & 0.9242$_{\pm 0.0347}$ \\
FacesUCR & 0.7141$_{\pm 0.003}$ & 0.7951$_{\pm 0.0034}$ & 0.8354$_{\pm 0.0054}$ & \textbf{0.9080$_{\pm 0.0025}$} \\
FiftyWords & 0.5949$_{\pm 0.0125}$ & 0.6777$_{\pm 0.0111}$ & 0.6462$_{\pm 0.0096}$ & \textbf{0.7604$_{\pm 0.0076}$} \\
Fish & 0.9162$_{\pm 0.0033}$ & 0.8705$_{\pm 0.0201}$ & \textbf{0.9333$_{\pm 0.0066}$} & 0.9010$_{\pm 0.0087}$ \\
FordA & 0.8932$_{\pm 0.002}$ & \textbf{0.9015$_{\pm 0.0008}$} & 0.8581$_{\pm 0.0048}$ & 0.8939$_{\pm 0.0023}$ \\
FordB & 0.7642$_{\pm 0.0119}$ & 0.765$_{\pm 0.0019}$ & 0.7305$_{\pm 0.0031}$ & \textbf{0.7790$_{\pm 0.0012}$} \\
FreezerRegularTrain & 0.9738$_{\pm 0.0013}$ & 0.8994$_{\pm 0.0012}$ & 0.9374$_{\pm 0.0043}$ & \textbf{0.9812$_{\pm 0.0041}$} \\
FreezerSmallTrain & \textbf{0.9549$_{\pm 0.0059}$} & 0.7758$_{\pm 0.0067}$ & 0.7942$_{\pm 0.0059}$ & 0.9421$_{\pm 0.0106}$ \\
Fungi & 0.7043$_{\pm 0.0093}$ & \textbf{0.9964$_{\pm 0.0062}$} & 0.8262$_{\pm 0.0164}$ & 0.9892$_{\pm 0.0000}$ \\
GestureMidAirD1 & 0.6744$_{\pm 0.0379}$ & 0.6744$_{\pm 0.0044}$ & 0.659$_{\pm 0.0044}$ & \textbf{0.7615$_{\pm 0.0077}$} \\
GestureMidAirD2 & 0.5692$_{\pm 0.0}$ & 0.5744$_{\pm 0.016}$ & 0.6154$_{\pm 0.0077}$ & \textbf{0.7103$_{\pm 0.0089}$} \\
GestureMidAirD3 & 0.3974$_{\pm 0.0247}$ & 0.359$_{\pm 0.0118}$ & 0.3282$_{\pm 0.016}$ & \textbf{0.4077$_{\pm 0.0077}$} \\
GesturePebbleZ1 & 0.8915$_{\pm 0.0067}$ & 0.8469$_{\pm 0.0067}$ & \textbf{0.9283$_{\pm 0.0034}$} & 0.8760$_{\pm 0.0089}$ \\
GesturePebbleZ2 & 0.8165$_{\pm 0.011}$ & 0.8376$_{\pm 0.0073}$ & \textbf{0.9219$_{\pm 0.0256}$} & 0.7932$_{\pm 0.0037}$ \\
GunPoint & 0.9444$_{\pm 0.0038}$ & \textbf{1.0$_{\pm 0.0}$} & 0.98$_{\pm 0.0067}$ & 0.9911$_{\pm 0.0038}$ \\
GunPointAgeSpan & 0.9684$_{\pm 0.0032}$ & 0.962$_{\pm 0.0032}$ & 0.9905$_{\pm 0.0}$ & \textbf{0.9926$_{\pm 0.0037}$} \\
GunPointMaleVersusFemale & 0.962$_{\pm 0.0032}$ & 0.9884$_{\pm 0.0018}$ & \textbf{0.9958$_{\pm 0.0018}$} & 0.9852$_{\pm 0.0018}$ \\
GunPointOldVersusYoung & \textbf{1.0$_{\pm 0.0}$} & 0.9577$_{\pm 0.008}$ & 0.9968$_{\pm 0.0}$ & 0.9937$_{\pm 0.0032}$ \\
Ham & 0.7206$_{\pm 0.022}$ & \textbf{0.7714$_{\pm 0.0}$} & 0.673$_{\pm 0.0145}$ & 0.7302$_{\pm 0.0198}$ \\
HandOutlines & 0.9045$_{\pm 0.0062}$ & 0.909$_{\pm 0.0056}$ & \textbf{0.9162$_{\pm 0.0072}$} & 0.8712$_{\pm 0.0165}$ \\
Haptics & 0.4481$_{\pm 0.0056}$ & \textbf{0.5152$_{\pm 0.0068}$} & 0.4968$_{\pm 0.0032}$ & 0.5076$_{\pm 0.0050}$ \\
Herring & 0.599$_{\pm 0.0325}$ & 0.6146$_{\pm 0.009}$ & \textbf{0.6667$_{\pm 0.0239}$} & \textbf{0.6667$_{\pm 0.0090}$} \\
HouseTwenty & 0.8655$_{\pm 0.0084}$ & 0.9356$_{\pm 0.0097}$ & 0.9412$_{\pm 0.0}$ & \textbf{0.9580$_{\pm 0.0000}$} \\
InlineSkate & 0.357$_{\pm 0.0105}$ & 0.3176$_{\pm 0.001}$ & \textbf{0.363$_{\pm 0.0136}$} & 0.3388$_{\pm 0.0038}$ \\
InsectEPGRegularTrain & \textbf{1.0$_{\pm 0.0}$} & 0.9183$_{\pm 0.0061}$ & \textbf{1.0$_{\pm 0.0}$} & \textbf{1.0000$_{\pm 0.0000}$} \\
InsectEPGSmallTrain & \textbf{1.0$_{\pm 0.0}$} & 0.8501$_{\pm 0.0061}$ & \textbf{1.0$_{\pm 0.0}$} & \textbf{1.0000$_{\pm 0.0000}$} \\
InsectWingbeatSound & 0.5066$_{\pm 0.0066}$ & \textbf{0.6158$_{\pm 0.0037}$} & 0.519$_{\pm 0.0044}$ & 0.5614$_{\pm 0.0024}$ \\
ItalyPowerDemand & 0.8698$_{\pm 0.007}$ & \textbf{0.9498$_{\pm 0.0006}$} & 0.9077$_{\pm 0.0035}$ & 0.9326$_{\pm 0.0050}$ \\
LargeKitchenAppliances & 0.7227$_{\pm 0.0027}$ & 0.7333$_{\pm 0.0071}$ & 0.7804$_{\pm 0.0041}$ & \textbf{0.8249$_{\pm 0.0086}$} \\
Lightning2 & 0.6885$_{\pm 0.0164}$ & 0.7377$_{\pm 0.0328}$ & 0.8033$_{\pm 0.0}$ & \textbf{0.8197$_{\pm 0.0164}$} \\
Lightning7 & 0.6895$_{\pm 0.0158}$ & 0.6758$_{\pm 0.0209}$ & 0.7763$_{\pm 0.0285}$ & \textbf{0.8493$_{\pm 0.0000}$} \\
Mallat & 0.8304$_{\pm 0.0025}$ & 0.859$_{\pm 0.0111}$ & 0.8903$_{\pm 0.0137}$ & \textbf{0.8948$_{\pm 0.0101}$} \\
Meat & 0.8889$_{\pm 0.0192}$ & \textbf{0.9444$_{\pm 0.0096}$} & 0.9389$_{\pm 0.0096}$ & \textbf{0.9444$_{\pm 0.0096}$} \\
MedicalImages & 0.7044$_{\pm 0.0027}$ & 0.6939$_{\pm 0.0062}$ & 0.7079$_{\pm 0.0035}$ & \textbf{0.7303$_{\pm 0.0039}$} \\
MelbournePedestrian & \textbf{0.912$_{\pm 0.0045}$} & 0.8662$_{\pm 0.0047}$ & 0.9016$_{\pm 0.0031}$ & 0.9077$_{\pm 0.0025}$ \\
MiddlePhalanxOutlineAgeGroup & 0.6169$_{\pm 0.0065}$ & 0.5779$_{\pm 0.0065}$ & 0.5801$_{\pm 0.0099}$ & \textbf{0.6494$_{\pm 0.0112}$} \\
MiddlePhalanxOutlineCorrect & 0.7858$_{\pm 0.0099}$ & \textbf{0.8625$_{\pm 0.006}$} & 0.8099$_{\pm 0.0099}$ & 0.8419$_{\pm 0.0060}$ \\
MiddlePhalanxTW & 0.5238$_{\pm 0.0037}$ & \textbf{0.5952$_{\pm 0.0037}$} & 0.5368$_{\pm 0.0099}$ & 0.5887$_{\pm 0.0246}$ \\
MixedShapesRegularTrain & 0.9392$_{\pm 0.0013}$ & 0.9124$_{\pm 0.0023}$ & \textbf{0.943$_{\pm 0.0044}$} & 0.9340$_{\pm 0.0004}$ \\
MixedShapesSmallTrain & \textbf{0.9061$_{\pm 0.0029}$} & 0.8389$_{\pm 0.0041}$ & 0.8961$_{\pm 0.0004}$ & 0.8841$_{\pm 0.0023}$ \\
MoteStrain & \textbf{0.9462$_{\pm 0.0032}$} & 0.8914$_{\pm 0.0083}$ & 0.9137$_{\pm 0.0136}$ & 0.9271$_{\pm 0.0039}$ \\
NonInvasiveFetalECGThorax1 & 0.7696$_{\pm 0.0049}$ & \textbf{0.8887$_{\pm 0.0035}$} & 0.6222$_{\pm 0.0037}$ & 0.8195$_{\pm 0.0075}$ \\
NonInvasiveFetalECGThorax2 & 0.811$_{\pm 0.0016}$ & \textbf{0.9138$_{\pm 0.0006}$} & 0.6872$_{\pm 0.0025}$ & 0.8422$_{\pm 0.0028}$ \\
OSULeaf & 0.7975$_{\pm 0.0072}$ & 0.7355$_{\pm 0.0041}$ & 0.8747$_{\pm 0.0104}$ & \textbf{0.8926$_{\pm 0.0072}$} \\
OliveOil & 0.7111$_{\pm 0.0192}$ & 0.9$_{\pm 0.0}$ & \textbf{0.9333$_{\pm 0.0}$} & 0.4000$_{\pm 0.0000}$ \\
PLAID & 0.7933$_{\pm 0.0113}$ & 0.7312$_{\pm 0.0088}$ & \textbf{0.8181$_{\pm 0.0047}$} & 0.7374$_{\pm 0.0085}$ \\
PhalangesOutlinesCorrect & 0.7743$_{\pm 0.0041}$ & \textbf{0.8248$_{\pm 0.0007}$} & 0.7786$_{\pm 0.0042}$ & 0.7541$_{\pm 0.0012}$ \\
Phoneme & 0.2802$_{\pm 0.0044}$ & 0.2751$_{\pm 0.0062}$ & \textbf{0.323$_{\pm 0.0034}$} & 0.3193$_{\pm 0.0032}$ \\
PickupGestureWiimoteZ & 0.6933$_{\pm 0.0808}$ & 0.68$_{\pm 0.06}$ & 0.74$_{\pm 0.02}$ & \textbf{0.8467$_{\pm 0.0115}$} \\
PigAirwayPressure & 0.3782$_{\pm 0.01}$ & 0.1186$_{\pm 0.0028}$ & \textbf{0.484$_{\pm 0.0147}$} & 0.3013$_{\pm 0.0147}$ \\
PigArtPressure & \textbf{0.9391$_{\pm 0.0028}$} & 0.6106$_{\pm 0.0048}$ & 0.9103$_{\pm 0.0028}$ & 0.7532$_{\pm 0.0121}$ \\
PigCVP & \textbf{0.8381$_{\pm 0.0242}$} & 0.609$_{\pm 0.0373}$ & 0.7837$_{\pm 0.0127}$ & 0.7596$_{\pm 0.0083}$ \\
Plane & 0.9937$_{\pm 0.0055}$ & 0.9841$_{\pm 0.0145}$ & \textbf{1.0$_{\pm 0.0}$} & \textbf{1.0000$_{\pm 0.0000}$} \\
PowerCons & 0.9352$_{\pm 0.0032}$ & \textbf{0.9463$_{\pm 0.0032}$} & 0.9093$_{\pm 0.0032}$ & 0.9222$_{\pm 0.0056}$ \\
ProximalPhalanxOutlineAgeGroup & 0.8537$_{\pm 0.0049}$ & 0.839$_{\pm 0.0049}$ & 0.8537$_{\pm 0.0049}$ & \textbf{0.8780$_{\pm 0.0000}$} \\
ProximalPhalanxOutlineCorrect & 0.8385$_{\pm 0.0034}$ & \textbf{0.8751$_{\pm 0.0052}$} & 0.8202$_{\pm 0.0086}$ & 0.7915$_{\pm 0.0130}$ \\
ProximalPhalanxTW & 0.8065$_{\pm 0.0056}$ & 0.8114$_{\pm 0.0028}$ & 0.7691$_{\pm 0.0028}$ & \textbf{0.8130$_{\pm 0.0028}$} \\
RefrigerationDevices & \textbf{0.5564$_{\pm 0.0041}$} & 0.536$_{\pm 0.0046}$ & 0.504$_{\pm 0.0122}$ & 0.5227$_{\pm 0.0027}$ \\
Rock & 0.6067$_{\pm 0.0306}$ & \textbf{0.84$_{\pm 0.02}$} & 0.7133$_{\pm 0.0115}$ & 0.6933$_{\pm 0.0115}$ \\
ScreenType & \textbf{0.5324$_{\pm 0.0077}$} & 0.4436$_{\pm 0.0178}$ & 0.4649$_{\pm 0.0134}$ & 0.5253$_{\pm 0.0053}$ \\
SemgHandGenderCh2 & 0.8928$_{\pm 0.0042}$ & 0.8028$_{\pm 0.0079}$ & \textbf{0.9189$_{\pm 0.0086}$} & 0.8739$_{\pm 0.0042}$ \\
SemgHandMovementCh2 & 0.72$_{\pm 0.0038}$ & 0.5237$_{\pm 0.0013}$ & \textbf{0.797$_{\pm 0.0056}$} & 0.6111$_{\pm 0.0022}$ \\
SemgHandSubjectCh2 & 0.7741$_{\pm 0.0134}$ & 0.6815$_{\pm 0.0026}$ & \textbf{0.8622$_{\pm 0.0135}$} & 0.7444$_{\pm 0.0059}$ \\
ShakeGestureWiimoteZ & \textbf{0.9267$_{\pm 0.0115}$} & 0.8533$_{\pm 0.0115}$ & 0.8867$_{\pm 0.0115}$ & 0.8467$_{\pm 0.0115}$ \\
ShapeletSim & 0.8833$_{\pm 0.0111}$ & 0.9648$_{\pm 0.0064}$ & 0.9278$_{\pm 0.0111}$ & \textbf{0.9741$_{\pm 0.0140}$} \\
ShapesAll & 0.8194$_{\pm 0.0025}$ & 0.8239$_{\pm 0.0092}$ & 0.8194$_{\pm 0.0054}$ & \textbf{0.8789$_{\pm 0.0025}$} \\
SmallKitchenAppliances & 0.8027$_{\pm 0.0071}$ & 0.72$_{\pm 0.0141}$ & 0.8089$_{\pm 0.0081}$ & \textbf{0.8436$_{\pm 0.0015}$} \\
SmoothSubspace & 0.8933$_{\pm 0.0067}$ & 0.92$_{\pm 0.0133}$ & 0.9067$_{\pm 0.0115}$ & \textbf{0.9311$_{\pm 0.0102}$} \\
SonyAIBORobotSurface1 & 0.7987$_{\pm 0.0044}$ & 0.8087$_{\pm 0.006}$ & 0.787$_{\pm 0.0294}$ & \textbf{0.8569$_{\pm 0.0152}$} \\
SonyAIBORobotSurface2 & 0.8297$_{\pm 0.0034}$ & 0.8353$_{\pm 0.0106}$ & \textbf{0.8552$_{\pm 0.0168}$} & 0.8475$_{\pm 0.0121}$ \\
StarLightCurves & \textbf{0.9792$_{\pm 0.0002}$} & 0.9734$_{\pm 0.0004}$ & 0.9759$_{\pm 0.0005}$ & 0.9779$_{\pm 0.0002}$ \\
Strawberry & 0.9378$_{\pm 0.0072}$ & \textbf{0.9604$_{\pm 0.0016}$} & 0.9514$_{\pm 0.0027}$ & 0.9144$_{\pm 0.0016}$ \\
SwedishLeaf & 0.9205$_{\pm 0.0065}$ & 0.9205$_{\pm 0.0051}$ & \textbf{0.9275$_{\pm 0.0009}$} & 0.9029$_{\pm 0.0040}$ \\
Symbols & 0.9407$_{\pm 0.0126}$ & 0.938$_{\pm 0.0126}$ & 0.9698$_{\pm 0.0056}$ & \textbf{0.9749$_{\pm 0.0017}$} \\
SyntheticControl & 0.9667$_{\pm 0.0033}$ & 0.9433$_{\pm 0.0}$ & 0.9767$_{\pm 0.0067}$ & \textbf{0.9900$_{\pm 0.0000}$} \\
ToeSegmentation1 & 0.8596$_{\pm 0.0044}$ & 0.924$_{\pm 0.0127}$ & \textbf{0.9635$_{\pm 0.0067}$} & 0.9532$_{\pm 0.0110}$ \\
ToeSegmentation2 & 0.7256$_{\pm 0.016}$ & 0.8462$_{\pm 0.0077}$ & \textbf{0.9282$_{\pm 0.0044}$} & 0.8923$_{\pm 0.0133}$ \\
Trace & \textbf{1.0$_{\pm 0.0}$} & 0.99$_{\pm 0.0}$ & \textbf{1.0$_{\pm 0.0}$} & \textbf{1.0000$_{\pm 0.0000}$} \\
TwoLeadECG & 0.8712$_{\pm 0.0303}$ & 0.9649$_{\pm 0.0116}$ & 0.9962$_{\pm 0.0005}$ & \textbf{0.9965$_{\pm 0.0009}$} \\
TwoPatterns & 0.8414$_{\pm 0.0041}$ & 0.8919$_{\pm 0.0079}$ & 0.8802$_{\pm 0.0079}$ & \textbf{0.9693$_{\pm 0.0005}$} \\
UMD & 0.9329$_{\pm 0.0223}$ & 0.9699$_{\pm 0.004}$ & 0.9722$_{\pm 0.0069}$ & \textbf{0.9931$_{\pm 0.0000}$} \\
UWaveGestureLibraryAll & 0.8773$_{\pm 0.0028}$ & 0.8975$_{\pm 0.0021}$ & 0.8458$_{\pm 0.0057}$ & \textbf{0.9181$_{\pm 0.0002}$} \\
UWaveGestureLibraryX & 0.8053$_{\pm 0.0007}$ & 0.7829$_{\pm 0.0071}$ & 0.7696$_{\pm 0.0028}$ & \textbf{0.8142$_{\pm 0.0009}$} \\
UWaveGestureLibraryY & \textbf{0.7338$_{\pm 0.0032}$} & 0.7022$_{\pm 0.0026}$ & 0.6874$_{\pm 0.0022}$ & 0.7261$_{\pm 0.0032}$ \\
UWaveGestureLibraryZ & 0.7427$_{\pm 0.0036}$ & 0.73$_{\pm 0.002}$ & 0.7324$_{\pm 0.0036}$ & \textbf{0.7639$_{\pm 0.0046}$} \\
Wafer & \textbf{0.9933$_{\pm 0.0006}$} & 0.9867$_{\pm 0.0003}$ & 0.9903$_{\pm 0.0003}$ & 0.9921$_{\pm 0.0011}$ \\
Wine & 0.7469$_{\pm 0.0107}$ & \textbf{0.8457$_{\pm 0.0283}$} & 0.7901$_{\pm 0.0107}$ & 0.5802$_{\pm 0.0107}$ \\
WordSynonyms & 0.5199$_{\pm 0.0127}$ & \textbf{0.6003$_{\pm 0.0078}$} & 0.5502$_{\pm 0.0081}$ & 0.5664$_{\pm 0.0065}$ \\
Worms & \textbf{0.7403$_{\pm 0.0}$} & 0.7056$_{\pm 0.015}$ & 0.6537$_{\pm 0.027}$ & 0.6450$_{\pm 0.0075}$ \\
WormsTwoClass & 0.7835$_{\pm 0.0075}$ & 0.7706$_{\pm 0.0075}$ & \textbf{0.8139$_{\pm 0.0198}$} & 0.7273$_{\pm 0.0225}$ \\
Yoga & 0.8219$_{\pm 0.0015}$ & \textbf{0.8264$_{\pm 0.003}$} & 0.815$_{\pm 0.0012}$ & 0.6860$_{\pm 0.0111}$ \\
\midrule
\# Wins & 18 & 34 & 33 & \textbf{52} \\
Avg.~Rank & 2.88 & 2.67 & 2.37 & \textbf{2.08} \\
Avg.~Acc. & 0.7714 & 0.7786 & 0.7922 & \textbf{0.7944} \\
\bottomrule
\end{longtable}

\begin{center}
\footnotesize
\begin{longtable}{l r r r r r}
\caption{Comparison under the fine tuning regime on UCR.
The best epoch performance is averaged over 3 random seeds and reported with the standard deviation.
Bold indicates the best result per row. GPT4S, NuTime, Moment and Mantis results are from \citep{feofanov2025mantislightweightcalibratedfoundation}.} \\
\toprule

 \textbf{Dataset} & \textbf{GPT4TS} & \textbf{NuTime} & \textbf{Moment} & \textbf{Mantis} & \textbf{Utica} \\ \midrule

\endfirsthead
\multicolumn{6}{c}{\tablename\ \thetable{} -- \textit{continued from previous page}} \\
\toprule
\textbf{Dataset} & \textbf{GPT4TS} & \textbf{NuTime} & \textbf{Moment} & \textbf{Mantis} & \textbf{Utica} \\
\midrule
\endhead
\midrule
\multicolumn{6}{r}{\textit{Continued on next page}} \\
\endfoot
\bottomrule
\endlastfoot

ACSF1 & 0.4900$_{\pm0.0265}$ & 0.6733$_{\pm0.0321}$ & 0.6033$_{\pm0.0379}$ & \textbf{0.7433$_{\pm0.0115}$} & 0.6733$_{\pm0.0419}$ \\
Adiac & 0.3572$_{\pm0.0059}$ & 0.7349$_{\pm0.0141}$ & 0.5823$_{\pm0.0296}$ & 0.7766$_{\pm0.0030}$ & \textbf{0.8159$_{\pm0.0130}$} \\
AllGestureWiimoteX & 0.4895$_{\pm0.0105}$ & 0.6386$_{\pm0.0038}$ & 0.7000$_{\pm0.0103}$ & \textbf{0.7619$_{\pm0.0050}$} & 0.7395$_{\pm0.0115}$ \\
AllGestureWiimoteY & 0.5062$_{\pm0.0116}$ & 0.7152$_{\pm0.0103}$ & 0.7081$_{\pm0.0343}$ & \textbf{0.7948$_{\pm0.0097}$} & 0.7390$_{\pm0.0106}$ \\
AllGestureWiimoteZ & 0.4719$_{\pm0.0119}$ & 0.6419$_{\pm0.0058}$ & 0.6976$_{\pm0.0218}$ & \textbf{0.7300$_{\pm0.0100}$} & 0.6967$_{\pm0.0141}$ \\
ArrowHead & 0.7638$_{\pm0.0033}$ & 0.8305$_{\pm0.0033}$ & 0.7790$_{\pm0.0682}$ & 0.8210$_{\pm0.0389}$ & \textbf{0.8457$_{\pm0.0123}$} \\
BME & 0.9444$_{\pm0.0102}$ & \textbf{1.0000$_{\pm0.0000}$} & 0.9800$_{\pm0.0231}$ & 0.9956$_{\pm0.0077}$ & \textbf{1.0000$_{\pm0.0000}$} \\
Beef & \textbf{0.8000$_{\pm0.0577}$} & \textbf{0.8222$_{\pm0.0192}$} & \textbf{0.8000$_{\pm0.0577}$} & 0.7000$_{\pm0.0333}$ & 0.7889$_{\pm0.0157}$ \\
BeetleFly & 0.8167$_{\pm0.0289}$ & 0.8833$_{\pm0.0764}$ & 0.8000$_{\pm0.1500}$ & 0.8833$_{\pm0.0764}$ & \textbf{0.9500$_{\pm0.0000}$} \\
BirdChicken & 0.6167$_{\pm0.0289}$ & 0.8500$_{\pm0.0500}$ & 0.7667$_{\pm0.0289}$ & \textbf{0.9000$_{\pm0.0500}$} & \textbf{0.9000$_{\pm0.0000}$} \\
CBF & 0.8596$_{\pm0.0083}$ & 0.9652$_{\pm0.0046}$ & 0.9767$_{\pm0.0172}$ & 0.9848$_{\pm0.0074}$ & \textbf{0.9993$_{\pm0.0005}$} \\
Car & 0.8000$_{\pm0.0167}$ & 0.8444$_{\pm0.0096}$ & \textbf{0.8778$_{\pm0.0419}$} & 0.8722$_{\pm0.0096}$ & 0.8389$_{\pm0.0079}$ \\
Chinatown & 0.9718$_{\pm0.0089}$ & 0.9738$_{\pm0.0000}$ & 0.9708$_{\pm0.0127}$ & 0.9718$_{\pm0.0017}$ & \textbf{0.9796$_{\pm0.0024}$} \\
ChlorineConcentration & 0.5733$_{\pm0.0091}$ & 0.6718$_{\pm0.0080}$ & 0.6289$_{\pm0.0178}$ & 0.6971$_{\pm0.0112}$ & \textbf{0.7218$_{\pm0.0145}$} \\
CinCECGTorso & 0.7930$_{\pm0.0353}$ & \textbf{0.9302$_{\pm0.0067}$} & 0.7430$_{\pm0.0080}$ & 0.7814$_{\pm0.0173}$ & 0.8517$_{\pm0.0188}$ \\
Coffee & \textbf{1.0000$_{\pm0.0000}$} & \textbf{1.0000$_{\pm0.0000}$} & 0.9881$_{\pm0.0206}$ & \textbf{1.0000$_{\pm0.0000}$} & \textbf{1.0000$_{\pm0.0000}$} \\
Computers & 0.6200$_{\pm0.0106}$ & \textbf{0.8120$_{\pm0.0174}$} & 0.6747$_{\pm0.0295}$ & 0.7813$_{\pm0.0162}$ & 0.7733$_{\pm0.0105}$ \\
CricketX & 0.5231$_{\pm0.0143}$ & 0.7453$_{\pm0.0039}$ & 0.7915$_{\pm0.0192}$ & 0.7966$_{\pm0.0090}$ & \textbf{0.8120$_{\pm0.0032}$} \\
CricketY & 0.5419$_{\pm0.0090}$ & 0.7479$_{\pm0.0104}$ & 0.7487$_{\pm0.0271}$ & 0.8060$_{\pm0.0141}$ & \textbf{0.8555$_{\pm0.0105}$} \\
CricketZ & 0.5496$_{\pm0.0039}$ & 0.7701$_{\pm0.0246}$ & 0.8068$_{\pm0.0218}$ & 0.8043$_{\pm0.0415}$ & \textbf{0.8350$_{\pm0.0134}$} \\
Crop & 0.7311$_{\pm0.0033}$ & 0.7574$_{\pm0.0022}$ & 0.7170$_{\pm0.0071}$ & 0.7444$_{\pm0.0039}$ & \textbf{0.7618$_{\pm0.0011}$} \\
DiatomSizeReduction & 0.9401$_{\pm0.0361}$ & 0.9314$_{\pm0.0425}$ & 0.9325$_{\pm0.0576}$ & \textbf{0.9684$_{\pm0.0136}$} & 0.9564$_{\pm0.0152}$ \\
DistalPhalanxOutlineAgeGroup & 0.7218$_{\pm0.0042}$ & 0.7386$_{\pm0.0150}$ & 0.7362$_{\pm0.0272}$ & 0.7698$_{\pm0.0259}$ & \textbf{0.7722$_{\pm0.0034}$} \\
DistalPhalanxOutlineCorrect & 0.6932$_{\pm0.0084}$ & 0.7705$_{\pm0.0302}$ & 0.7452$_{\pm0.0200}$ & 0.7717$_{\pm0.0158}$ & \textbf{0.7995$_{\pm0.0062}$} \\
DistalPhalanxTW & 0.6978$_{\pm0.0125}$ & 0.6978$_{\pm0.0216}$ & 0.6451$_{\pm0.0300}$ & 0.6954$_{\pm0.0231}$ & \textbf{0.7218$_{\pm0.0136}$} \\
DodgerLoopDay & 0.5542$_{\pm0.0564}$ & 0.5458$_{\pm0.0688}$ & 0.5292$_{\pm0.0641}$ & 0.6250$_{\pm0.0250}$ & \textbf{0.6625$_{\pm0.0102}$} \\
DodgerLoopGame & 0.8551$_{\pm0.0192}$ & 0.8213$_{\pm0.0649}$ & 0.8599$_{\pm0.0233}$ & \textbf{0.8841$_{\pm0.0145}$} & 0.8575$_{\pm0.0068}$ \\
DodgerLoopWeekend & \textbf{0.9855$_{\pm0.0000}$} & 0.9783$_{\pm0.0126}$ & 0.9783$_{\pm0.0126}$ & 0.9783$_{\pm0.0000}$ & \textbf{0.9855$_{\pm0.0000}$} \\
ECG200 & 0.8467$_{\pm0.0551}$ & 0.8700$_{\pm0.0100}$ & \textbf{0.8867$_{\pm0.0231}$} & 0.8567$_{\pm0.0058}$ & 0.8833$_{\pm0.0094}$ \\
ECG5000 & \textbf{0.9408$_{\pm0.0019}$} & 0.9379$_{\pm0.0001}$ & 0.9325$_{\pm0.0003}$ & 0.9335$_{\pm0.0081}$ & 0.9403$_{\pm0.0007}$ \\
ECGFiveDays & 0.8784$_{\pm0.0760}$ & \textbf{0.9621$_{\pm0.0084}$} & 0.7944$_{\pm0.0373}$ & 0.9148$_{\pm0.0292}$ & 0.9241$_{\pm0.0054}$ \\
Earthquakes & 0.7338$_{\pm0.0072}$ & 0.7482$_{\pm0.0000}$ & \textbf{0.7578$_{\pm0.0166}$} & 0.7482$_{\pm0.0000}$ & 0.7506$_{\pm0.0034}$ \\
ElectricDevices & 0.5776$_{\pm0.0173}$ & 0.7112$_{\pm0.0109}$ & 0.6666$_{\pm0.0204}$ & 0.7454$_{\pm0.0049}$ & \textbf{0.7700$_{\pm0.0013}$} \\
FaceAll & 0.7247$_{\pm0.0042}$ & \textbf{0.8385$_{\pm0.0098}$} & 0.8012$_{\pm0.0080}$ & 0.8308$_{\pm0.0031}$ & 0.8008$_{\pm0.0254}$ \\
FaceFour & 0.8295$_{\pm0.0301}$ & 0.9318$_{\pm0.0227}$ & 0.8371$_{\pm0.0853}$ & \textbf{0.9773$_{\pm0.0114}$} & 0.8977$_{\pm0.0093}$ \\
FacesUCR & 0.7932$_{\pm0.0047}$ & 0.8846$_{\pm0.0127}$ & 0.8350$_{\pm0.0014}$ & 0.9148$_{\pm0.0071}$ & \textbf{0.9260$_{\pm0.0039}$} \\
FiftyWords & 0.6615$_{\pm0.0058}$ & 0.7875$_{\pm0.0083}$ & 0.7861$_{\pm0.0222}$ & 0.8139$_{\pm0.0083}$ & \textbf{0.8411$_{\pm0.0089}$} \\
Fish & 0.8133$_{\pm0.0119}$ & 0.9524$_{\pm0.0119}$ & 0.9162$_{\pm0.0201}$ & \textbf{0.9714$_{\pm0.0099}$} & 0.9486$_{\pm0.0093}$ \\
FordA & 0.7207$_{\pm0.0798}$ & 0.9225$_{\pm0.0095}$ & 0.9202$_{\pm0.0271}$ & 0.9343$_{\pm0.0046}$ & \textbf{0.9391$_{\pm0.0019}$} \\
FordB & 0.6107$_{\pm0.0567}$ & 0.7922$_{\pm0.0126}$ & 0.8173$_{\pm0.0125}$ & 0.7979$_{\pm0.0135}$ & \textbf{0.8297$_{\pm0.0046}$} \\
FreezerRegularTrain & 0.8884$_{\pm0.0329}$ & 0.9942$_{\pm0.0014}$ & 0.9876$_{\pm0.0056}$ & 0.9927$_{\pm0.0041}$ & \textbf{0.9967$_{\pm0.0012}$} \\
FreezerSmallTrain & 0.6614$_{\pm0.0094}$ & 0.9537$_{\pm0.0290}$ & 0.8206$_{\pm0.0695}$ & \textbf{0.9667$_{\pm0.0098}$} & 0.9213$_{\pm0.0121}$ \\
Fungi & 0.7061$_{\pm0.0124}$ & 0.7294$_{\pm0.0135}$ & 0.6900$_{\pm0.0350}$ & 0.7778$_{\pm0.0031}$ & \textbf{0.7867$_{\pm0.0134}$} \\
GestureMidAirD1 & 0.5615$_{\pm0.0353}$ & 0.7410$_{\pm0.0222}$ & 0.6487$_{\pm0.0311}$ & 0.7692$_{\pm0.0407}$ & \textbf{0.7872$_{\pm0.0158}$} \\
GestureMidAirD2 & 0.4846$_{\pm0.0428}$ & 0.6513$_{\pm0.0347}$ & 0.5615$_{\pm0.0077}$ & 0.6744$_{\pm0.0118}$ & \textbf{0.7359$_{\pm0.0130}$} \\
GestureMidAirD3 & 0.3308$_{\pm0.0400}$ & 0.4359$_{\pm0.0089}$ & 0.3615$_{\pm0.0277}$ & 0.4077$_{\pm0.0231}$ & \textbf{0.4897$_{\pm0.0221}$} \\
GesturePebbleZ1 & 0.8217$_{\pm0.0034}$ & 0.8953$_{\pm0.0058}$ & 0.9109$_{\pm0.0089}$ & 0.9322$_{\pm0.0089}$ & \textbf{0.9477$_{\pm0.0047}$} \\
GesturePebbleZ2 & 0.7553$_{\pm0.0159}$ & 0.8861$_{\pm0.0219}$ & 0.8861$_{\pm0.0063}$ & \textbf{0.9241$_{\pm0.0219}$} & 0.8628$_{\pm0.0196}$ \\
GunPoint & 0.8733$_{\pm0.0067}$ & 0.9933$_{\pm0.0000}$ & 0.9867$_{\pm0.0115}$ & \textbf{1.0000$_{\pm0.0000}$} & 0.9933$_{\pm0.0000}$ \\
GunPointAgeSpan & 0.8755$_{\pm0.0174}$ & 0.9916$_{\pm0.0018}$ & 0.9662$_{\pm0.0048}$ & \textbf{0.9979$_{\pm0.0037}$} & 0.9958$_{\pm0.0015}$ \\
GunPointMaleVersusFemale & 0.9610$_{\pm0.0391}$ & 0.9989$_{\pm0.0018}$ & 0.9916$_{\pm0.0048}$ & \textbf{1.0000$_{\pm0.0000}$} & 0.9937$_{\pm0.0000}$ \\
GunPointOldVersusYoung & \textbf{1.0000$_{\pm0.0000}$} & \textbf{1.0000$_{\pm0.0000}$} & 0.9820$_{\pm0.0048}$ & \textbf{1.0000$_{\pm0.0000}$} & 0.9947$_{\pm0.0054}$ \\
Ham & 0.6698$_{\pm0.0240}$ & 0.7238$_{\pm0.0190}$ & \textbf{0.7746$_{\pm0.0306}$} & 0.7238$_{\pm0.0415}$ & 0.7302$_{\pm0.0119}$ \\
Haptics & 0.4426$_{\pm0.0167}$ & 0.4870$_{\pm0.0142}$ & 0.4167$_{\pm0.0179}$ & \textbf{0.5227$_{\pm0.0149}$} & 0.5217$_{\pm0.0110}$ \\
Herring & 0.6458$_{\pm0.0325}$ & 0.6458$_{\pm0.0393}$ & 0.5677$_{\pm0.0325}$ & 0.6354$_{\pm0.0325}$ & \textbf{0.6510$_{\pm0.0266}$} \\
HouseTwenty & 0.8123$_{\pm0.0049}$ & 0.9272$_{\pm0.0097}$ & 0.8992$_{\pm0.0084}$ & 0.9804$_{\pm0.0049}$ & \textbf{0.9832$_{\pm0.0000}$} \\
InsectEPGRegularTrain & \textbf{1.0000$_{\pm0.0000}$} & \textbf{1.0000$_{\pm0.0000}$} & 0.9331$_{\pm0.0267}$ & \textbf{1.0000$_{\pm0.0000}$} & \textbf{1.0000$_{\pm0.0000}$} \\
InsectEPGSmallTrain & \textbf{1.0000$_{\pm0.0000}$} & \textbf{1.0000$_{\pm0.0000}$} & 0.8969$_{\pm0.0023}$ & \textbf{1.0000$_{\pm0.0000}$} & \textbf{1.0000$_{\pm0.0000}$} \\
InsectWingbeatSound & 0.6165$_{\pm0.0069}$ & 0.6190$_{\pm0.0048}$ & 0.6019$_{\pm0.0034}$ & 0.5961$_{\pm0.0051}$ & \textbf{0.6328$_{\pm0.0075}$} \\
ItalyPowerDemand & 0.9637$_{\pm0.0011}$ & \textbf{0.9666$_{\pm0.0020}$} & 0.9624$_{\pm0.0006}$ & 0.9624$_{\pm0.0015}$ & 0.9543$_{\pm0.0016}$ \\
LargeKitchenAppliances & 0.5156$_{\pm0.0285}$ & 0.8240$_{\pm0.0116}$ & 0.8693$_{\pm0.0232}$ & 0.8516$_{\pm0.0111}$ & \textbf{0.8951$_{\pm0.0063}$} \\
Lightning2 & 0.7268$_{\pm0.0250}$ & 0.7760$_{\pm0.0250}$ & 0.7978$_{\pm0.0250}$ & 0.8087$_{\pm0.0189}$ & \textbf{0.8580$_{\pm0.0077}$} \\
Lightning7 & 0.5982$_{\pm0.0079}$ & 0.7032$_{\pm0.0519}$ & 0.7717$_{\pm0.0209}$ & 0.7397$_{\pm0.0362}$ & \textbf{0.8310$_{\pm0.0065}$} \\
Mallat & 0.9200$_{\pm0.0258}$ & 0.9508$_{\pm0.0191}$ & 0.9205$_{\pm0.0260}$ & 0.9404$_{\pm0.0072}$ & \textbf{0.9598$_{\pm0.0064}$} \\
Meat & 0.6944$_{\pm0.2269}$ & \textbf{0.9611$_{\pm0.0255}$} & 0.6444$_{\pm0.0674}$ & 0.9333$_{\pm0.0289}$ & 0.9222$_{\pm0.0157}$ \\
MedicalImages & 0.6110$_{\pm0.0201}$ & 0.7513$_{\pm0.0149}$ & 0.7206$_{\pm0.0290}$ & 0.7662$_{\pm0.0178}$ & \textbf{0.7943$_{\pm0.0075}$} \\
MelbournePedestrian & 0.9351$_{\pm0.0037}$ & 0.9601$_{\pm0.0019}$ & 0.8795$_{\pm0.0055}$ & 0.9552$_{\pm0.0010}$ & \textbf{0.9692$_{\pm0.0015}$} \\
MiddlePhalanxOutlineAgeGroup & 0.6104$_{\pm0.0065}$ & 0.5368$_{\pm0.0198}$ & 0.6126$_{\pm0.0327}$ & 0.5996$_{\pm0.0209}$ & \textbf{0.6580$_{\pm0.0031}$} \\
MiddlePhalanxOutlineCorrect & 0.6518$_{\pm0.0436}$ & 0.7938$_{\pm0.0248}$ & 0.8076$_{\pm0.0034}$ & 0.8339$_{\pm0.0139}$ & \textbf{0.8625$_{\pm0.0028}$} \\
MiddlePhalanxTW & \textbf{0.5909$_{\pm0.0000}$} & 0.4632$_{\pm0.0150}$ & 0.5606$_{\pm0.0037}$ & 0.4827$_{\pm0.0300}$ & 0.5866$_{\pm0.0061}$ \\
MixedShapesSmallTrain & 0.8115$_{\pm0.0111}$ & 0.9281$_{\pm0.0054}$ & 0.8357$_{\pm0.0363}$ & \textbf{0.9531$_{\pm0.0038}$} & 0.9434$_{\pm0.0037}$ \\
MoteStrain & 0.8243$_{\pm0.0062}$ & \textbf{0.9609$_{\pm0.0037}$} & 0.8341$_{\pm0.0264}$ & 0.9068$_{\pm0.0174}$ & 0.9297$_{\pm0.0078}$ \\
NonInvasiveFetalECGThorax1 & 0.7774$_{\pm0.0152}$ & \textbf{0.9342$_{\pm0.0076}$} & 0.7535$_{\pm0.0237}$ & 0.9086$_{\pm0.0063}$ & 0.9318$_{\pm0.0054}$ \\
NonInvasiveFetalECGThorax2 & 0.8390$_{\pm0.0098}$ & 0.9381$_{\pm0.0046}$ & 0.7995$_{\pm0.0033}$ & 0.9216$_{\pm0.0055}$ & \textbf{0.9454$_{\pm0.0036}$} \\
OSULeaf & 0.4669$_{\pm0.0180}$ & 0.8967$_{\pm0.0286}$ & 0.8017$_{\pm0.0219}$ & \textbf{0.9642$_{\pm0.0024}$} & 0.9229$_{\pm0.0186}$ \\
OliveOil & 0.4889$_{\pm0.0385}$ & 0.7778$_{\pm0.0385}$ & 0.4667$_{\pm0.0667}$ & \textbf{0.8889$_{\pm0.0509}$} & 0.6778$_{\pm0.0956}$ \\
PhalangesOutlinesCorrect & 0.6437$_{\pm0.0041}$ & 0.8263$_{\pm0.0031}$ & 0.8030$_{\pm0.0047}$ & 0.8162$_{\pm0.0128}$ & \textbf{0.8400$_{\pm0.0038}$} \\
Phoneme & 0.1466$_{\pm0.0023}$ & 0.2973$_{\pm0.0081}$ & 0.2904$_{\pm0.0090}$ & \textbf{0.3214$_{\pm0.0059}$} & 0.3087$_{\pm0.0051}$ \\
PickupGestureWiimoteZ & 0.7133$_{\pm0.0231}$ & 0.7267$_{\pm0.0115}$ & 0.7200$_{\pm0.0000}$ & \textbf{0.7600$_{\pm0.0529}$} & 0.7400$_{\pm0.0000}$ \\
Plane & 0.9778$_{\pm0.0055}$ & \textbf{1.0000$_{\pm0.0000}$} & 0.9873$_{\pm0.0145}$ & \textbf{1.0000$_{\pm0.0000}$} & \textbf{1.0000$_{\pm0.0000}$} \\
PowerCons & \textbf{1.0000$_{\pm0.0000}$} & 0.9889$_{\pm0.0111}$ & 0.9167$_{\pm0.0111}$ & 0.9944$_{\pm0.0096}$ & 0.9926$_{\pm0.0026}$ \\
ProximalPhalanxOutlineAgeGroup & 0.8016$_{\pm0.0203}$ & 0.8553$_{\pm0.0102}$ & 0.8455$_{\pm0.0123}$ & 0.8455$_{\pm0.0246}$ & \textbf{0.8780$_{\pm0.0040}$} \\
ProximalPhalanxOutlineCorrect & 0.8179$_{\pm0.0034}$ & \textbf{0.9072$_{\pm0.0060}$} & 0.8259$_{\pm0.0221}$ & 0.8763$_{\pm0.0209}$ & 0.8923$_{\pm0.0113}$ \\
ProximalPhalanxTW & 0.7967$_{\pm0.0102}$ & 0.8179$_{\pm0.0056}$ & 0.7870$_{\pm0.0102}$ & 0.7772$_{\pm0.0314}$ & \textbf{0.8212$_{\pm0.0083}$} \\
RefrigerationDevices & 0.4453$_{\pm0.0092}$ & \textbf{0.5511$_{\pm0.0227}$} & 0.4996$_{\pm0.0336}$ & 0.4916$_{\pm0.0111}$ & 0.5396$_{\pm0.0163}$ \\
Rock & 0.5933$_{\pm0.0503}$ & 0.6800$_{\pm0.0200}$ & 0.6733$_{\pm0.0611}$ & \textbf{0.7133$_{\pm0.0231}$} & 0.5667$_{\pm0.0189}$ \\
ScreenType & 0.3956$_{\pm0.0126}$ & 0.5289$_{\pm0.0434}$ & 0.4213$_{\pm0.0116}$ & 0.4782$_{\pm0.0041}$ & \textbf{0.5520$_{\pm0.0058}$} \\
ShakeGestureWiimoteZ & 0.6867$_{\pm0.0115}$ & 0.9133$_{\pm0.0231}$ & 0.9067$_{\pm0.0115}$ & \textbf{0.9200$_{\pm0.0000}$} & 0.8267$_{\pm0.0340}$ \\
ShapeletSim & 0.5074$_{\pm0.0128}$ & 0.8852$_{\pm0.0534}$ & \textbf{1.0000$_{\pm0.0000}$} & \textbf{1.0000$_{\pm0.0000}$} & \textbf{1.0000$_{\pm0.0000}$} \\
ShapesAll & 0.6833$_{\pm0.0029}$ & 0.8856$_{\pm0.0067}$ & 0.8039$_{\pm0.0108}$ & 0.8906$_{\pm0.0059}$ & \textbf{0.9067$_{\pm0.0085}$} \\
SmallKitchenAppliances & 0.5529$_{\pm0.0171}$ & 0.7822$_{\pm0.0015}$ & 0.6756$_{\pm0.0161}$ & 0.7929$_{\pm0.0031}$ & \textbf{0.8516$_{\pm0.0063}$} \\
SmoothSubspace & 0.8911$_{\pm0.0038}$ & 0.9889$_{\pm0.0038}$ & 0.7733$_{\pm0.0291}$ & \textbf{0.9978$_{\pm0.0038}$} & 0.9867$_{\pm0.0054}$ \\
SonyAIBORobotSurface1 & 0.6267$_{\pm0.0215}$ & 0.8525$_{\pm0.0189}$ & 0.7604$_{\pm0.0204}$ & 0.8142$_{\pm0.0113}$ & \textbf{0.9567$_{\pm0.0151}$} \\
SonyAIBORobotSurface2 & 0.8454$_{\pm0.0205}$ & 0.8506$_{\pm0.0115}$ & 0.8765$_{\pm0.0276}$ & 0.8996$_{\pm0.0112}$ & \textbf{0.9266$_{\pm0.0082}$} \\
Strawberry & 0.7261$_{\pm0.0788}$ & \textbf{0.9712$_{\pm0.0031}$} & 0.9459$_{\pm0.0054}$ & 0.9667$_{\pm0.0078}$ & 0.9595$_{\pm0.0117}$ \\
SwedishLeaf & 0.8203$_{\pm0.0305}$ & 0.9573$_{\pm0.0024}$ & 0.9376$_{\pm0.0070}$ & \textbf{0.9712$_{\pm0.0042}$} & 0.9504$_{\pm0.0114}$ \\
Symbols & 0.8003$_{\pm0.0243}$ & 0.9779$_{\pm0.0053}$ & 0.9481$_{\pm0.0192}$ & \textbf{0.9889$_{\pm0.0017}$} & 0.9689$_{\pm0.0079}$ \\
SyntheticControl & 0.9622$_{\pm0.0117}$ & 0.9889$_{\pm0.0051}$ & 0.9889$_{\pm0.0019}$ & \textbf{0.9944$_{\pm0.0038}$} & 0.9900$_{\pm0.0000}$ \\
ToeSegmentation1 & 0.5687$_{\pm0.0051}$ & 0.9459$_{\pm0.0101}$ & 0.9284$_{\pm0.0483}$ & 0.9664$_{\pm0.0067}$ & \textbf{0.9752$_{\pm0.0021}$} \\
ToeSegmentation2 & 0.6923$_{\pm0.0353}$ & 0.8974$_{\pm0.0089}$ & \textbf{0.9308$_{\pm0.0204}$} & 0.9231$_{\pm0.0077}$ & 0.9000$_{\pm0.0218}$ \\
Trace & 0.8567$_{\pm0.0577}$ & \textbf{1.0000$_{\pm0.0000}$} & \textbf{1.0000$_{\pm0.0000}$} & \textbf{1.0000$_{\pm0.0000}$} & \textbf{1.0000$_{\pm0.0000}$} \\
TwoLeadECG & 0.7934$_{\pm0.0272}$ & 0.9104$_{\pm0.0273}$ & 0.8879$_{\pm0.0509}$ & \textbf{0.9941$_{\pm0.0018}$} & 0.9936$_{\pm0.0042}$ \\
TwoPatterns & 0.9676$_{\pm0.0173}$ & 0.9998$_{\pm0.0001}$ & 0.9998$_{\pm0.0002}$ & 0.9999$_{\pm0.0001}$ & \textbf{1.0000$_{\pm0.0000}$} \\
UMD & 0.9699$_{\pm0.0080}$ & \textbf{1.0000$_{\pm0.0000}$} & 0.9815$_{\pm0.0200}$ & 0.9907$_{\pm0.0040}$ & 0.9977$_{\pm0.0033}$ \\
UWaveGestureLibraryX & 0.7319$_{\pm0.0069}$ & 0.8516$_{\pm0.0044}$ & 0.8062$_{\pm0.0076}$ & \textbf{0.8679$_{\pm0.0028}$} & 0.8655$_{\pm0.0033}$ \\
UWaveGestureLibraryY & 0.6145$_{\pm0.0027}$ & 0.7701$_{\pm0.0050}$ & 0.7378$_{\pm0.0127}$ & 0.7917$_{\pm0.0073}$ & \textbf{0.7930$_{\pm0.0021}$} \\
UWaveGestureLibraryZ & 0.6224$_{\pm0.0217}$ & 0.7813$_{\pm0.0073}$ & 0.7570$_{\pm0.0158}$ & 0.8053$_{\pm0.0011}$ & \textbf{0.8119$_{\pm0.0040}$} \\
Wafer & 0.9948$_{\pm0.0022}$ & 0.9987$_{\pm0.0003}$ & 0.9982$_{\pm0.0004}$ & \textbf{0.9991$_{\pm0.0003}$} & 0.9981$_{\pm0.0003}$ \\
Wine & 0.5185$_{\pm0.0321}$ & \textbf{0.8025$_{\pm0.0466}$} & 0.5000$_{\pm0.0000}$ & 0.6605$_{\pm0.0107}$ & 0.5864$_{\pm0.0231}$ \\
WordSynonyms & 0.5397$_{\pm0.0096}$ & 0.6917$_{\pm0.0065}$ & 0.7079$_{\pm0.0220}$ & \textbf{0.7429$_{\pm0.0031}$} & 0.7132$_{\pm0.0201}$ \\
Worms & 0.4416$_{\pm0.0225}$ & 0.7229$_{\pm0.0198}$ & 0.7013$_{\pm0.0225}$ & 0.7143$_{\pm0.0260}$ & \textbf{0.7489$_{\pm0.0324}$} \\
WormsTwoClass & 0.6364$_{\pm0.0130}$ & 0.7143$_{\pm0.0225}$ & 0.7532$_{\pm0.0450}$ & \textbf{0.8052$_{\pm0.0344}$} & 0.7922$_{\pm0.0106}$ \\
Yoga & 0.7467$_{\pm0.0025}$ & 0.8053$_{\pm0.0118}$ & 0.8477$_{\pm0.0057}$ & 0.8709$_{\pm0.0132}$ & \textbf{0.8775$_{\pm0.0062}$} \\
\midrule
\textbf{\# Wins} & 8 & 21 & 7 & 38 & \textbf{60} \\
\textbf{Avg. Acc.} & 0.7434 & 0.8387 & 0.7991 & 0.8504 & \textbf{0.8569} \\
\textbf{Avg. Rank} & 4.4248 & 2.2301 & 3.5133 & 1.9469 & \textbf{1.4071} \\

\end{longtable}
\end{center}

\end{document}